\documentclass{article}

\usepackage[nonatbib, final]{score_workshop}

\pdfoutput=1
\usepackage[utf8]{inputenc} % allow utf-8 input
\usepackage[T1]{fontenc}    % use 8-bit T1 fonts
\usepackage{hyperref}       % hyperlinks
\usepackage{url}            % simple URL typesetting
\usepackage{booktabs}       % professional-quality tables
\usepackage{amsfonts}       % blackboard math symbols
\usepackage{nicefrac}       % compact symbols for 1/2, etc.
\usepackage{microtype}      % microtypography
\usepackage{xcolor}         % colors

\usepackage{graphicx}
\usepackage{caption}
\usepackage{subcaption}
\usepackage{amsmath}
\usepackage{amssymb}
\usepackage{cleveref}
\usepackage{cancel}
\usepackage{multirow}
\usepackage{wrapfig}

\allowdisplaybreaks

\usepackage{empheq}
\newcommand*\widefbox[1]{\fbox{\hspace{2em}#1\hspace{2em}}}

\usepackage{amsmath,amsfonts,bm}

\def\eqref#1{equation~\ref{#1}}
\def\1{\bm{1}}

\def\eps{{\epsilon}}

\def\rvepsilon{{\bm{\epsilon}}}
\def\rvtheta{{\bm{\theta}}}
\def\rva{{\mathbf{a}}}
\def\rvb{{\mathbf{b}}}

\def\rvf{{\mathbf{f}}}
\def\rvg{{\mathbf{g}}}

\def\rvu{{\mathbf{i}}}

\def\rvs{{\mathbf{s}}}

\def\rvu{{\mathbf{u}}}
\def\rvv{{\mathbf{v}}}
\def\rvw{{\mathbf{w}}}
\def\rvx{{\mathbf{x}}}
\def\rvy{{\mathbf{y}}}
\def\rvz{{\mathbf{z}}}

\def\rmA{{\mathbf{A}}}

\def\rmC{{\mathbf{C}}}

\def\rmG{{\mathbf{G}}}

\def\rmI{{\mathbf{I}}}

\def\rmK{{\mathbf{K}}}
\def\rmL{{\mathbf{L}}}

\def\rmP{{\mathbf{P}}}
\def\rmQ{{\mathbf{Q}}}

\def\rmW{{\mathbf{W}}}

\def\vzero{{\bm{0}}}

\def\vmu{{\bm{\mu}}}

\def\evtheta{{\theta}}

\def\mK{{\bm{K}}}

\def\mLambda{{\bm{\Lambda}}}
\def\mSigma{{\bm{\Sigma}}}

\DeclareMathAlphabet{\mathsfit}{\encodingdefault}{\sfdefault}{m}{sl}
\SetMathAlphabet{\mathsfit}{bold}{\encodingdefault}{\sfdefault}{bx}{n}

\def\gC{{\mathcal{C}}}

\def\gG{{\mathcal{G}}}

\def\gI{{\mathcal{I}}}

\def\gL{{\mathcal{L}}}

\def\gN{{\mathcal{N}}}

\def\gU{{\mathcal{U}}}

\newcommand{\Var}{\mathrm{Var}}

\usepackage{url}
\usepackage{color}
\usepackage{graphicx}
\usepackage{float}
\usepackage{subcaption}
\usepackage{multirow}
\usepackage{siunitx}
\newcommand{\Norm}[1]{\left\Vert#1\right\Vert}
\newcommand{\abs}[1]{\vert#1\vert}
\newcommand{\Abs}[1]{\left\vert#1\right\vert}

\newcommand{\bq}{\begin{equation}}
\newcommand{\eq}{\end{equation}}

\newcommand{\loss}{\ell}

\newcommand{\D}{{\mathrm{d}}}
\newcommand{\EE}{\mathbb{E}}

\newcommand{\balpha}{\bar\alpha}

\newcommand\numberthis{\addtocounter{equation}{1}\tag{\theequation}}

\newcommand{\x}{{\mkern-2mu\times\mkern-2mu}}

\newcommand\conj{H}
\newcommand\trns{T}
\newcommand{\Real}{\text{Real}}
\newcommand{\real}{\text{real}}

\allowdisplaybreaks

\newcommand{\ud}{\mathrm{d}}

\newcommand{\SMLD}{\text{SMLD}}
\newcommand{\DDPM}{\text{DDPM}}
\newcommand{\GFF}{\text{GFF}}
\newcommand{\NI}{\text{NI}}
\newcommand{\GFFDDPM}{\operatorname{NI-DDPM}}

\usepackage{xcolor}
\title{Score-based Denoising Diffusion with \\ Non-Isotropic Gaussian Noise Models}

\author{%
    Vikram Voleti \\
    Mila, University of Montreal\\
    Canada
   \And
   Adam Oberman \\
   Mila, McGill University \\
   Canada
   \And
   Christopher Pal \\
   Mila, Polytechnique Montr\'eal \\
   CIFAR AI Chair, Service Now\\
   Canada
}

\begin{document}

\maketitle

\begin{abstract}
Generative models based on denoising diffusion techniques have led to an unprecedented increase in the quality and diversity of imagery that is now possible to create with neural generative models. However, most contemporary state-of-the-art methods are derived from a standard isotropic Gaussian formulation. In this work we examine the situation where non-isotropic Gaussian distributions are used. We present the key mathematical derivations for creating denoising diffusion models using an underlying non-isotropic Gaussian noise model. We also provide initial experiments with the CIFAR10 dataset to help verify empirically that this more general modelling approach can also yield high-quality samples.
\end{abstract}

\section{Introduction}

\begin{wrapfigure}{R}{0.38\textwidth}
\vspace{-.5cm}
\centering
\begin{subfigure}{.18\textwidth}
  \centering
  \includegraphics[width=\linewidth]{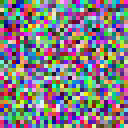}
  \caption{Isotropic}
  \label{fig:sub1}
\end{subfigure}
\begin{subfigure}{.18\textwidth}
  \centering
  \includegraphics[width=\linewidth]{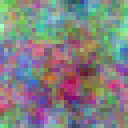}
  \caption{Non-isotropic}
  \label{fig:sub2}
\end{subfigure}
\vspace{-.2cm}
\caption{Gaussian noise samples.
}
\vspace{-.5cm}
\label{fig:gff_sample}
\end{wrapfigure}
Score-based denoising diffusion models~\cite{song2019generative, ho2020ddpm, song2020sde} have seen great success as generative models for images~\cite{dhariwal2021diffusion, song2020improved}, as well as other modes such as video~\cite{ho2022VDM, yang2022ResidualVideoDiffusion, voleti2022MCVD}, audio~\cite{kong2021diffwave, chen2021wavegrad}, etc. The underlying framework relies on a noising "forward" process that adds noise to real images (or other data), and a denoising "reverse" process that iteratively removes noise. In most cases, the noise distribution used is the isotropic Gaussian i.e. noise samples are independently and identically distributed (IID) as the standard normal at each pixel.

In this work, we lay the theoretical foundations and derive the key mathematics for a non-isotropic Gaussian formulation for denoising diffusion models. It is our hope that these insights may open the door to new classes of models.
%This theory would be helpful for future work,,,,,..... 
% The motivation is that images have spatial correlation among its pixels, hence we consider noise images that also exhibit spatial correlations among their pixels, but randomly sampled from a noise distribution. 
One type of non-isotropic Gaussian noise arises in a family of models known as  Gaussian Free Fields (GFFs)~\cite{sheffield2007gff, berestycki2015gff, bramson2016gff, werner2020gff} (a.k.a. Gaussian Random Fields). GFF noise can be obtained by either convolving isotropic Gaussian noise with a filter, or applying frequency masking of noise.
% in the frequency space representation of that noise.
In either case this procedure allows one to model or generate smoother and correlated types of Gaussian noise.
In \Cref{fig:gff_sample,fig:GFFs}, we compare examples of isotropic Gaussian noise with GFF noise obtained using a frequency space window function consisting of $w(f)=\frac{1}{f}$.

Our contributions here consist of the following: (1) deriving the key mathematics for score-based denoising diffusion models using non-isotropic multivariate Gaussian distributions, (2) examining the special case of a GFF and the corresponding non-Isotropic Gaussian noise model, and (3) showing that diffusion models trained (eg. on the CIFAR-10 dataset~\cite{krizhevsky2009learning}) using a GFF noise process are also capable of yielding high-quality samples comparable to models based on isotropic Gaussian noise.

\Cref{sec:ddpm} and \Cref{sec:niddpm} contain more detailed derivations of the above equations for DDPM~\cite{ho2020ddpm} and our NI-DDPM. See \Cref{sec:smld} and \Cref{sec:nismld} for the equivalent derivations for Score Matching Langevin Dynamics (SMLD)~\cite{song2019generative,song2020improved}, and our Non-Isotropic SMLD (NI-SMLD).

\section{Isotropic Gaussian denoising diffusion models}

We perform our analysis below within the Denoising Diffusion Probabilistic Models (DDPM)~\cite{ho2020ddpm} framework, but our analysis is valid for all other types of score-based denoising diffusion models.

In DDPM, for a fixed sequence of positive scales $0 < \beta_1 < \cdots < \beta_L < 1$, $\balpha_t = \prod_{s=1}^t (1 - \beta_s)$, and a noise sample $\rvepsilon \sim \gN(\vzero, \rmI)$, the cumulative ``\textbf{forward}'' noising process is:
\begin{align}
% q_{t} (\rvx_t \mid \rvx_{t-1}) &= \gN(\rvx_t \mid \sqrt{1 - \beta_t} \rvx_{t-1}, \beta_t \rmI)
% \implies
% \rvx_t = \sqrt{1 - \beta_t}\rvx_{t-1} + \sqrt{\beta_t} \rvz_t \\
&q_t (\rvx_t \mid \rvx_0) = \gN(\sqrt{\balpha_t} \rvx_0, (1 - \balpha_t) \rmI) \implies \rvx_t = \sqrt{\balpha_t}\rvx_0 + \sqrt{1 - \balpha_t} \rvepsilon
\label{eq:DDPM_noise1}
% \\
% \implies &\log q (\rvx_t \mid \rvx_0) = \log C - \frac{1}{2(1 - \balpha_t)}(\rvx_t - \sqrt{\balpha_t}\rvx_0)^\trns(\rvx_t - \sqrt{\balpha_t}\rvx_0)
% \\
% \implies \text{Score }\approx\ &\nabla_{\rvx_t} \log q_{\balpha_t} (\rvx_t \mid \rvx_0) = -\frac{1}{(1 - \balpha_t)}(\rvx_t - \sqrt{\balpha_t} \rvx_0) = -\frac{1}{\sqrt{1 - \balpha_t}}\rvepsilon
\end{align}
The ``\textbf{reverse}'' process involves iteratively \textbf{sampling} $\rvx_{t-1}$ from $\rvx_t$ conditioned on $\rvx_0$ i.e. $p_{t-1}(\rvx_{t-1} \mid \rvx_t, \rvx_0)$, obtained from $q_t(\rvx_t \mid \rvx_0)$ using Bayes' rule. For this, first $\rvepsilon$ is estimated using a neural network $\rvepsilon_\rvtheta(\rvx_t, t)$. Then, using $\hat\rvx_0 = \big(\rvx_t - \sqrt{1 - \balpha_t}\rvepsilon_{\rvtheta}(\rvx_t, t)\big)/\sqrt{\balpha_t}$ from \cref{eq:DDPM_noise1}, $\rvx_{t-1}$ is sampled:
\begin{gather*}
p_{t-1}(\rvx_{t-1} \mid \rvx_t, \hat{\rvx}_0) = \gN(\ \tilde\vmu_{t}(\rvx_t, \hat{\rvx}_0), \tilde\beta_{t}\rmI \ ) \implies \rvx_{t-1} = \tilde\vmu_{t}(\rvx_t, \hat{\rvx}_0) + \sqrt{\tilde\beta_{t}}\rvz_t \quad \text{; where} \numberthis \\
\tilde\vmu_{t}(\rvx_t, \hat{\rvx}_0) = \frac{\sqrt{\balpha_{t-1}}\beta_t}{1 - \balpha_t}\hat{\rvx}_0 + \frac{\sqrt{1 - \beta_t}(1 - \balpha_{t-1})}{1 - \balpha_t}\rvx_t \ ;\ \tilde\beta_t = \frac{1 - \balpha_{t-1}}{1 - \balpha_t}\beta_t \ ;\ \rvz_t \sim \gN(\vzero, \rmI) \
\numberthis
\label{eq:reverse}
\end{gather*}
% \newpage
The\textbf{ objective function} to train $\rvepsilon_\rvtheta(\rvx_t, t)$ is simply an expected reconstruction loss with the true $\rvepsilon$:
\begin{align}
\gL_\rvepsilon(\rvtheta) = \EE_{t \sim \gU({1,\cdots,L}), \rvx_0 \sim p(\rvx_0), \rvepsilon \sim \gN(\vzero, \rmI)} \left[ \Norm{ \rvepsilon - \rvepsilon_\rvtheta\big(\sqrt{\balpha_t}\rvx_0 + \sqrt{1 - \balpha_t} \rvepsilon, t \big)}_2^2 \right]
\label{eq:eps_loss}
\end{align}
From the perspective of score matching, the \textbf{score} of the DDPM forward process is:
\begin{align}
\text{Score }\ \rvs = \ &\nabla_{\rvx_t} \log q_{t} (\rvx_t \mid \rvx_0) = -\frac{1}{(1 - \balpha_t)}(\rvx_t - \sqrt{\balpha_t} \rvx_0) = -\frac{1}{\sqrt{1 - \balpha_t}}\rvepsilon
\label{eq:score}
\end{align}
Thus, the overall \textbf{score-matching objective} for a score estimation network $\rvs_\rvtheta(\rvx_t, t)$ is the weighted sum of the loss $\loss_\rvs(\rvtheta; t)$ for each $t$, the weight being the inverse of the score \textbf{variance} at $t$ i.e. $(1 - \balpha_t)$:
\begin{align}
&\gL_\rvs(\rvtheta) = \EE_t\ (1 - \balpha_t)\ \loss_\rvs(\rvtheta; t) = \EE_{t, \rvx_0, \rvepsilon} \bigg[ \Norm{\sqrt{1 - \balpha_t}\rvs_\rvtheta (\rvx_t, t) + \rvepsilon }_2^2 \bigg]
\end{align}
When the score network output is redefined as per the \textbf{score-noise relationship} in \cref{eq:score}:
\begin{align}
\rvs_\rvtheta(\rvx_t, t) = -\frac{1}{\sqrt{1 - \balpha_t}}\rvepsilon_\evtheta(\rvx_t, t)\ \implies\ \gL_\rvs(\rvtheta) = \EE_{t, \rvx_0, \rvepsilon} \bigg[ \Norm{-\rvepsilon_\rvtheta (\rvx_t, t) + \rvepsilon }_2^2 \bigg] \quad = \gL_\rvepsilon(\rvtheta)
\end{align}
% $\rvs_\rvtheta(\rvx_t, t) = -\frac{1}{\sqrt{1 - \balpha_t}}\rvepsilon_\evtheta(\rvx_t, t)$,
% \begin{align}
% \rvs_\rvtheta(\rvx_t, t) = -\frac{1}{\sqrt{1 - \balpha_t}}\rvepsilon_\evtheta(\rvx_t, t),
% \end{align}
Thus, $\gL_\rvs = \gL_\rvepsilon$ i.e. the score-matching and noise reconstruction objectives are equivalent.

From \cite{saremi2019neb}, the \textbf{Expected Denoised Sample} (EDS) $\rvx_0^*(\rvx_t, t) \triangleq \EE_{\rvx_0 \sim p_{t}(\rvx_0 \mid \rvx_t)}[\rvx_0]$ and the score $\rvs$, estimated optimally as $\rvs_{\rvtheta^*}$, are related as:
\begin{align*}
&\rvs_{\rvtheta^*}(\rvx_t, t) = \EE\left[ \Norm{ \nabla_{\rvx_t} \log q_{t} (\rvx_t \mid \rvx_0) }_2^2 \right] \big(\rvx_0^* (\rvx_t, t) - \rvx_t\big) = \frac{1}{1 - \balpha_t}\big(\rvx_0^* (\rvx_t, t) - \rvx_t\big) \numberthis \\
\implies
&\rvx_0^*(\rvx_t, t) = \rvx_t + (1 - \balpha_t)\ \rvs_{\rvtheta^*}(\rvx_t, t) = \rvx_t - \sqrt{1 - \balpha_t} \ \rvepsilon_{\rvtheta^*}(\rvx_t, t)
\numberthis
\end{align*}
The EDS is often used to further improve the quality of the final image at $t=0$.

\section{Non-isotropic Gaussian denoising diffusion models}
We formulate the Non-Isotropic DDPM (\textbf{NI-DDPM}) using a non-isotropic Gaussian noise distribution with a positive semi-definite covariance matrix $\mSigma$ in the place of $\rmI$. The \textbf{forward} noising process is:
\begin{align}
% &q_t (\rvx_t \mid \rvx_{t-1}) = \gN(\sqrt{1 - \beta_t} \rvx_{t-1}, \beta_t \mSigma)
% \implies \rvx_t = \sqrt{1 - \beta_t}\rvx_{t-1} + \sqrt{\beta_t} \sqrt{\mSigma} \rvz_{t-1} \\
&q_t (\rvx_t \mid \rvx_0) = \gN(\sqrt{\balpha_t} \rvx_0, (1 - \balpha_t) \mSigma)
\implies \rvx_t = \sqrt{\balpha_t}\rvx_0 + \sqrt{1 - \balpha_t} \sqrt{\mSigma} \rvepsilon
\label{eq:niddpm_noise}
\end{align}
Thus, the \textbf{score} of NI-DDPM is (see \Cref{sec:niddpm_score} for derivation):
\begin{align}
&\text{Score }\ \rvs = \nabla_{\rvx_t} \log q_t (\rvx_t \mid \rvx_0) = -\mSigma^{-1}\frac{\rvx_t - \sqrt{\balpha_t} \rvx_0}{1 - \balpha_t} = -\frac{1}{\sqrt{1 - \balpha_t}}\sqrt{\mSigma^{-1}}\rvepsilon
\label{eq:GFF_score}
\end{align}
The \textbf{score-matching objective} for a score estimation network $\rvs_\rvtheta(\rvx_t, t)$ at each noise level $t$ is now:
\begin{align}
&\loss(\rvtheta; t) = \EE_{\rvx_0 \sim p(\rvx_0), \rvepsilon \sim \gN(\vzero, \rmI)} \bigg[ \Norm{\rvs_\rvtheta (\sqrt{\balpha_t}\rvx_0 + \sqrt{1 - \balpha_t} \rvepsilon, t) + \frac{1}{\sqrt{1 - \balpha_t}}\sqrt{\mSigma^{-1}}\rvepsilon }_2^2 \bigg]
\label{eq:GFF_score_loss_t}
\end{align}
The \textbf{variance} of this score is:
\begin{align*}
&\EE\left[ \Norm{ \nabla_{\rvx_t} \log q_{t}(\rvx_t \mid \rvx_0) }_2^2 \right]
= \EE\left[ \Norm{ -\frac{1}{\sqrt{1 - \balpha_t}}\sqrt{\mSigma^{-1}}\rvepsilon }_2^2 \right]
= \frac{1}{1 - \balpha_t}\mSigma^{-1}\EE\left[ \Norm{\rvepsilon}_2^2 \right]
\label{eq:GFF_variance}
\numberthis
\end{align*}
The \textbf{overall objective} is a weighted sum, the weight being the inverse of the score variance $(1 - \balpha_t)\mSigma$:
\begin{align}
\gL(\rvtheta) &= \EE_{t \sim \gU({1,\cdots,L})}\ (1 - \balpha_t)\mSigma \ \loss(\rvtheta; t) = \EE_{t, \rvx_0, \rvepsilon} \bigg[ \Norm{\sqrt{1 - \balpha_t}\sqrt{\mSigma}\rvs_\rvtheta (\rvx_t, t) + \rvepsilon }_2^2 \bigg]
\end{align}
Following the \textbf{score-noise relationship} in \cref{eq:GFF_score}:
\begin{align}
&\rvs_\rvtheta(\rvx_t, t) = -\frac{1}{\sqrt{1 - \balpha_t}}\sqrt{\mSigma^{-1}}\rvepsilon_\rvtheta(\rvx_t, t)
\label{eq:niddpm_stheta}
\end{align}
The \textbf{objective function} now becomes (expanding $\rvs_\rvtheta$ as per \cref{eq:niddpm_stheta}):
\begin{align}
&\gL(\rvtheta) = \EE_{t \sim \gU({1,\cdots,L}), \rvx_0 \sim p(\rvx_0), \rvepsilon \sim \gN(\vzero, \rmI)} \bigg[ \Norm{-\rvepsilon_\rvtheta (\sqrt{\balpha_t}\rvx_0 + \sqrt{1 - \balpha_t} \sqrt{\mSigma} \rvepsilon, t) + \rvepsilon }_2^2 \bigg]
\end{align}
This objective function for NI-DDPM seems like $\gL_\epsilon$ of DDPM, but DDPM's $\rvepsilon_\rvtheta$ network cannot be re-used here since their forward processes are different. DDPM produces $\rvx_t$ from $\rvx_0$ using \cref{eq:DDPM_noise1}, while NI-DDPM uses \cref{eq:niddpm_noise}. See \Cref{subsec:niddpm_objective} for alternate formulations of the score network.

\textbf{Sampling} involves computing $p_{t-1}(\rvx_{t-1} \mid \rvx_t, \hat{\rvx}_0)$ (see \Cref{subsec:sampling_niddpm} for derivation):
% \newpage
\begin{align*}
q_{t} (\rvx_t \mid \rvx_0) = \gN(\sqrt{\balpha_t} \rvx_0, (1 - \balpha_t) \mSigma)
&\implies \hat\rvx_0 = \frac{1}{\sqrt{\balpha_t}}\big(\rvx_t - \sqrt{1 - \balpha_t}\sqrt{\mSigma}\rvepsilon_{\rvtheta}(\rvx_t, t)\big)
\label{eq:niddpm_x0hat}
\numberthis
\\
p_{t-1}(\rvx_{t-1} \mid \rvx_t, \hat{\rvx}_0) = \gN( \tilde\vmu_t(\rvx_t, \hat{\rvx}_0), \tilde\beta_t\mSigma)
&\implies \rvx_{t-1}
= \tilde\vmu_t(\rvx_t, \hat{\rvx}_0) + \sqrt{\tilde\beta_t} \sqrt{\mSigma} \rvz_t
\numberthis
\end{align*}
where $\tilde\vmu_t$, $\tilde\beta_t$ and $\rvz_t$ are the same as \cref{eq:reverse}.

Alternatively, \cite{song2020sde} mentions using $\beta_t$ instead of $\tilde\beta_t$:
\begin{align*}
p_{t-1}^{\beta_t}(\rvx_{t-1} \mid \rvx_t, \hat{\rvx}_0) = \gN( \tilde\vmu_t(\rvx_t, \hat{\rvx}_0), \beta_t\mSigma)
&\implies \rvx_{t-1}
= \tilde\vmu_t(\rvx_t, \hat{\rvx}_0) + \sqrt{\beta_t} \sqrt{\mSigma} \rvz_t
\numberthis
\end{align*}
Alternatively, sampling using \textbf{DDIM}~\cite{song2020ddim} invokes the following distribution for $\rvx_{t-1}$:
\begin{align*}
% q_{t}(\rvx_L \mid \rvx_0) &= \gN(\rvx_L \mid \sqrt{\balpha_L}\rvx_0, (1 - \balpha_L)\mSigma)
% \numberthis
% \\
&p_{t-1}^{\text{DDIM}}(\rvx_{t-1} \mid \rvx_t, \hat \rvx_0) = \gN\left( \sqrt{\balpha_{t-1}}\hat{\rvx}_0 + \sqrt{1 - \balpha_{t-1}}\frac{\rvx_t - \sqrt{\balpha_t}\hat{\rvx}_0}{\sqrt{1 - \balpha_t}}, \ \ \vzero \right)
\numberthis
\\
% \implies q_{t}(\rvx_{t} \mid \rvx_0) &= \gN\left(\sqrt{\balpha_{t}}\rvx_0, (1 - \balpha_{t})\mSigma \right)
% \numberthis
% \\
% \implies \hat\rvx_0 &= \frac{1}{\sqrt{\balpha_t}}(\rvx_t - \sqrt{1 - \balpha_t}\sqrt{\mSigma}\rvz_{\rvtheta^*}^{(1)}(\rvx_t))
% \\
\implies
&\rvx_{t-1} = \sqrt{\balpha_{t-1}} \hat\rvx_0 + \sqrt{1 - \balpha_{t-1}} \sqrt{\mSigma} \rvepsilon_{\rvtheta}(\rvx_t, t)
\numberthis
\end{align*}
The \textbf{Expected Denoised Sample} $\rvx_0^*(\rvx_t, t)$ and the optimal score $\rvs_{\rvtheta^*}$ are now related as:
\begin{align*}
&\rvs_{\rvtheta^*}(\rvx_t, t) = \EE\left[ \Norm{ \nabla_{\rvx_t} \log q_{t} (\rvx_t \mid \rvx_0) }_2^2 \right] \big(\rvx_0^* (\rvx_t, t) - \rvx_t\big) = \frac{1}{1 - \balpha_t} \mSigma^{-1} \big(\rvx_0^* (\rvx_t, t) - \rvx_t\big) \numberthis \\
&\implies
\rvx_0^*(\rvx_t, t) = \rvx_t + (1 - \balpha_t)\ \mSigma \rvs_{\rvtheta^*}(\rvx_t, t) = \rvx_t - \sqrt{1 - \balpha_t} \ \sqrt{\mSigma} \rvepsilon_{\rvtheta^*}(\rvx_t, t)
\numberthis
\end{align*}

\textbf{SDE formulation}: Score-based diffusion models have also been analyzed as stochastic differential equations (SDEs)~\cite{song2020sde}. The SDE version of NI-DDPM, which we call Non-Isotropic Variance Preserving (NIVP) SDE, is (see \Cref{subsec:niddpm_sde} for derivation):
\begin{align*}
&\D \rvx = -\frac{1}{2}\beta(t)\rvx\ \D t + \sqrt{\beta(t)}\sqrt{\mSigma}\ \D \rvw
\numberthis
\\
\implies &p_{0t}\big(\rvx(t) \mid \rvx(0)\big) = \gN \left( \rvx(0)\ e^{-\frac{1}{2}\int_0^t \beta(s) \D s},\ \mSigma(\rmI - \rmI e^{-\int_0^t \beta(s) \D s}) \right)
\numberthis
\end{align*}

Finally, \Cref{sec:ddpm} and \Cref{sec:niddpm} contain more detailed derivations of the above equations for DDPM~\cite{ho2020ddpm} and our NI-DDPM. See \Cref{sec:smld} and \Cref{sec:nismld} for the equivalent derivations for Score Matching Langevin Dynamics (SMLD)~\cite{song2019generative,song2020improved}, and our Non-Isotropic SMLD (NI-SMLD).

\section{Gaussian Free Field (GFF) images}

A GFF image $\rvg$ can be obtained from a normal noise image $\rvz$ as follows~\cite{sheffield2007gff} (see \Cref{GFF} for more details):
\begin{enumerate}
\item
First, sample an $n \x n$ noise image $\rvz$ from the standard complex normal distribution 
with covariance matrix $\Gamma = \rmI_N$ where $N=n^2$ is the total number of pixels, and pseudo-covariance matrix $C = \vzero$
: $\rvz \sim \gC \gN(\vzero, \rmI_N, \vzero)$. (In principle, real noise could be used.)
\item
Apply the Discrete Fourier Transform using its $N \x N$ weights matrix $\rmW_N$: $\rmW_N \rvz$.
\item
Consider a diagonal $N \x N$ matrix of the reciprocal of an index value $|k_{ij}|$ per pixel $(i,j)$ in Fourier space : $\rmK^{-1} = [1/|k_{ij}|]_{(i,j)}$, and multiply this with the above: $\rmK^{-1} \rmW_N \rvz$.
\item
Take its Inverse Discrete Fourier Transform ($\rmW_N^{-1}$) to make the raw GFF image: $\rmW_N^{-1} \rmK^{-1} \rmW_N \rvz$.
However, this results in a GFF image with a small non-unit variance.
\item
Normalize the above GFF image with the standard deviation $\sigma_N$ at its resolution $N$, so that it has unit variance (see \Cref{subsec:gff_prob} for derivation of $\sigma_N$):
$\rvg_\text{complex} = \frac{1}{\sigma_N}\rmW_N^{-1} \rmK^{-1} \rmW_N \rvz$
\item
Extract only the real part of $\rvg_\text{complex}$, and normalize (see \Cref{subsec:GFF_real} for derivation):
\begin{align*}
\rvg &= \frac{1}{\sqrt{2N}\sigma_N} \Real \left(\rmW_N^{-1} \rmK^{-1} \rmW_N \rvz \right)
\numberthis
\label{eq:g}
\end{align*}
\end{enumerate}

See \Cref{fig:gff_sample,fig:GFFs} for examples of GFF images. Effectively, this procedure prioritizes lower frequencies over higher frequencies, thereby making the noise smoother, and hence correlated. The probability distribution of GFF images $\rvg$ can be seen as a non-isotropic multivariate Gaussian with mean $\vzero$, and a non-diagonal covariance matrix $\mSigma$ (see \Cref{subsec:gff_prob,subsec:GFF_real} for derivation):
\begin{align*}
p(\rvg) = \gN(\vzero, \mSigma); \mSigma = \sqrt{\mSigma} \sqrt{\mSigma}^\trns; \sqrt{\mSigma} &= \frac{1}{\sqrt{2N}\sigma_N}\Real\left(\rmW_N^{-1} \rmK^{-1} \rmW_N\right) \implies \rvg = \sqrt{\mSigma}\rvz
\numberthis
\label{eq:covG2}
% \\
% \sqrt{\mSigma^{-1}} &= \sigma_N \rmW_N^{-1} \rmK \rmW_N ; \quad
% \mSigma^{-1} = \sqrt{\mSigma^{-1}}^\trns \sqrt{\mSigma^{-1}},
% \numberthis
% \label{eq:covGinv2}
\end{align*}

% The log probability of the transformation from isotropic noise sample $\rvz$ to GFF noise sample $\rvg$ is (see \Cref{subsec:gff_logprob} for derivation):
% \begin{align*}
% \log p(\rvg) &= 
% % \log p(\rvz) - \log \Abs{\det \frac{\D \rvg}{\D \rvz}}
% % = \log p(\rvz) - \log \Abs{\det \sqrt{\mSigma}}
% % = \log p(\rvz) - \log \Abs{\det \frac{1}{\sqrt{2N}\sigma_N}\Real\left(\rmW_N^{-1} \rmK^{-1} \rmW_N\right) } 
% % = 
% \log p(\rvz) - \frac{1}{\sqrt{2N}\sigma_N} \log \Abs{\det\rmK^{-1} }
% \numberthis
% \end{align*}
% This is useful for formulating (normalizing) flows with non-isotropic Gaussian prior.

% \sloppy{Without the correcting factor of $\sigma_N$, the standard deviation of the resulting $\rvg_\real$ with are quite low
% % : for $2\x2, 4\x4, 8\x8, 16\x16, 32\x32, 64\x64, 128\x128$ are \allowbreak
% % % $0.39528470752104744, 0.17030533460540237, 0.053976619959778704, \allowbreak 0.015784561822114067, 0.004442407825782781, 0.0012219552979003889, \allowbreak 0.00033098344651100604$.
% % $0.3953, 0.1703, 0.0540, 0.0158, 0.0044, 0.0012, 0.0003$.
% }
% .
% To fix this, $\sigma_N$ equal to these values is divided.

\section{Results}

% \usepackage{multirow}
% \begin{table}[!htb]
\begin{wraptable}{r}{0.5\linewidth}
\vspace{-1.3em}
\caption{Image generation metrics FID, Precision (P), and Recall (R) for CIFAR10 using DDPM and NI-DDPM, with different generation steps.\vspace{-1.5em}}
\centering
\begin{tabular}{lrrcc}
% \hline
% \multicolumn{1}{|l|}{\textbf{10k samples}}                    & \multicolumn{1}{l|}{\textbf{steps}} & \multicolumn{1}{l|}{\textbf{FID}} & \multicolumn{1}{l|}{\textbf{Precision}} & \multicolumn{1}{l|}{\textbf{Recall}} \\ \hline
% \multicolumn{1}{|l|}{BASELINE\_DDPM}                          & \multicolumn{1}{r|}{1000}           & \multicolumn{1}{r|}{8.06}         & \multicolumn{1}{r|}{0.66}               & \multicolumn{1}{r|}{0.56}            \\ \hline
% \multicolumn{1}{|l|}{BASELINE\_DDPM\_HPmine}                  & \multicolumn{1}{r|}{1000}           & \multicolumn{1}{r|}{11.25}        & \multicolumn{1}{r|}{0.61}               & \multicolumn{1}{r|}{0.59}            \\ \hline
% \multicolumn{1}{|l|}{GFF\_DDPM}                               & \multicolumn{1}{r|}{1000}           & \multicolumn{1}{r|}{10.22}        & \multicolumn{1}{r|}{0.62}               & \multicolumn{1}{r|}{0.55}            \\ \hline
% \multicolumn{1}{|l|}{GFF\_DDPM\_HPmine}                       & \multicolumn{1}{r|}{1000}           & \multicolumn{1}{r|}{9.45}         & \multicolumn{1}{r|}{0.63}               & \multicolumn{1}{r|}{0.57}            \\ \hline
% &
\multicolumn{1}{l}{}                & \multicolumn{1}{l}{}              & \multicolumn{1}{l}{}                    & \multicolumn{1}{l}{}                 \\ \hline
\multicolumn{1}{|l|}{CIFAR10}                    & \multicolumn{1}{l|}{\textbf{steps}} & \multicolumn{1}{l|}{\textbf{FID} $\downarrow$} & \multicolumn{1}{l|}{\textbf{P} $\uparrow$} & \multicolumn{1}{l|}{\textbf{R} $\uparrow$} \\ \hline
\multicolumn{1}{|l|}{\multirow{5}{*}{DDPM}}         & \multicolumn{1}{r|}{1000}           & \multicolumn{1}{r|}{6.05}         & \multicolumn{1}{r|}{0.66}               & \multicolumn{1}{r|}{0.54}            \\ \cline{2-5} 
\multicolumn{1}{|l|}{}                                        & \multicolumn{1}{r|}{100}            & \multicolumn{1}{r|}{12.25}        & \multicolumn{1}{r|}{0.62}               & \multicolumn{1}{r|}{0.48}            \\ \cline{2-5} 
\multicolumn{1}{|l|}{}                                        & \multicolumn{1}{r|}{50}             & \multicolumn{1}{r|}{16.61}        & \multicolumn{1}{r|}{0.60}               & \multicolumn{1}{r|}{0.43}            \\ \cline{2-5} 
\multicolumn{1}{|l|}{}                                        & \multicolumn{1}{r|}{20}             & \multicolumn{1}{r|}{26.35}        & \multicolumn{1}{r|}{0.56}               & \multicolumn{1}{r|}{0.24}            \\ \cline{2-5} 
\multicolumn{1}{|l|}{}                                        & \multicolumn{1}{r|}{10}             & \multicolumn{1}{r|}{44.95}        & \multicolumn{1}{r|}{0.49}               & \multicolumn{1}{r|}{0.24}            \\ \hline
% \multicolumn{1}{|l|}{\multirow{5}{*}{BASELINE\_DDPM\_HPmine}} & \multicolumn{1}{r|}{1000}           & \multicolumn{1}{r|}{9.11}         & \multicolumn{1}{r|}{0.61}               & \multicolumn{1}{r|}{0.56}            \\ \cline{2-5} 
% \multicolumn{1}{|l|}{}                                        & \multicolumn{1}{r|}{100}            & \multicolumn{1}{r|}{15.20}        & \multicolumn{1}{r|}{0.58}               & \multicolumn{1}{r|}{0.50}            \\ \cline{2-5} 
% \multicolumn{1}{|l|}{}                                        & \multicolumn{1}{r|}{50}             & \multicolumn{1}{r|}{20.25}        & \multicolumn{1}{r|}{0.56}               & \multicolumn{1}{r|}{0.47}            \\ \cline{2-5} 
% \multicolumn{1}{|l|}{}                                        & \multicolumn{1}{r|}{20}             & \multicolumn{1}{r|}{31.88}        & \multicolumn{1}{r|}{0.52}               & \multicolumn{1}{r|}{0.38}            \\ \cline{2-5} 
% \multicolumn{1}{|l|}{}                                        & \multicolumn{1}{r|}{10}             & \multicolumn{1}{r|}{49.78}        & \multicolumn{1}{r|}{0.45}               & \multicolumn{1}{r|}{0.30}            \\ \hline
\multicolumn{1}{|l|}{\multirow{5}{*}{NI-DDPM}}              & \multicolumn{1}{r|}{1000}           & \multicolumn{1}{r|}{6.95}         & \multicolumn{1}{r|}{0.62}               & \multicolumn{1}{r|}{0.53}            \\ \cline{2-5} 
\multicolumn{1}{|l|}{}                                        & \multicolumn{1}{r|}{100}            & \multicolumn{1}{r|}{12.68}        & \multicolumn{1}{r|}{0.60}               & \multicolumn{1}{r|}{0.49}            \\ \cline{2-5} 
\multicolumn{1}{|l|}{}                                        & \multicolumn{1}{r|}{50}             & \multicolumn{1}{r|}{16.91}        & \multicolumn{1}{r|}{0.57}               & \multicolumn{1}{r|}{0.45}            \\ \cline{2-5} 
\multicolumn{1}{|l|}{}                                        & \multicolumn{1}{r|}{20}             & \multicolumn{1}{r|}{30.41}        & \multicolumn{1}{r|}{0.52}               & \multicolumn{1}{r|}{0.35}            \\ \cline{2-5} 
\multicolumn{1}{|l|}{}                                        & \multicolumn{1}{r|}{10}             & \multicolumn{1}{r|}{60.32}        & \multicolumn{1}{r|}{0.43}               & \multicolumn{1}{r|}{0.23}            \\ \hline
% \multicolumn{1}{|l|}{\multirow{5}{*}{GFF\_DDPM\_HPmine}}      & \multicolumn{1}{r|}{1000}           & \multicolumn{1}{r|}{7.24}         & \multicolumn{1}{r|}{0.62}               & \multicolumn{1}{r|}{0.54}            \\ \cline{2-5} 
% \multicolumn{1}{|l|}{}                                        & \multicolumn{1}{r|}{100}            & \multicolumn{1}{r|}{11.47}        & \multicolumn{1}{r|}{0.59}               & \multicolumn{1}{r|}{0.51}            \\ \cline{2-5} 
% \multicolumn{1}{|l|}{}                                        & \multicolumn{1}{r|}{50}             & \multicolumn{1}{r|}{16.17}        & \multicolumn{1}{r|}{0.57}               & \multicolumn{1}{r|}{0.47}            \\ \cline{2-5} 
% \multicolumn{1}{|l|}{}                                        & \multicolumn{1}{r|}{20}             & \multicolumn{1}{r|}{32.12}        & \multicolumn{1}{r|}{0.51}               & \multicolumn{1}{r|}{0.38}            \\ \cline{2-5} 
% \multicolumn{1}{|l|}{}                                        & \multicolumn{1}{r|}{10}             & \multicolumn{1}{r|}{59.23}        & \multicolumn{1}{r|}{0.44}               & \multicolumn{1}{r|}{0.27}            \\ \hline
\end{tabular}
\label{tab:1}
\vspace{-1em}
\end{wraptable}
We train two models on CIFAR10, one using DDPM and the other using NI-DDPM with the exact same hyperparameters (batch size, learning rate, etc.) for 300,000 iterations. We then sample 50,000 images from each, and calculate the image generation metrics of Fr\'echet Inception Distance (FID)~\cite{heusel2017gans}, Precision (P), and Recall (R).
% For FID, lower is better, while for Precision and Recall, higher is better.
Although the models were trained on 1000 steps between data and noise, we report these metrics while sampling images using 1000, and smaller steps: 100, 50, 20, 10.

As can be seen from \Cref{tab:1}, our non-isotropic variant performs comparable to the isotropic baseline. The difference between them increases with decreasing number of steps between noise and data. This provides a reasonable proof-of-concept that non-isotropic Gaussian noise works just as well as isotropic noise when used in denoising diffusion models for image generation.

\section{Conclusion}
We have presented the key mathematics behind non-isotropic Gaussian DDPMs, as well as a complete example using a GFF. We then noted quantitative comparison of using GFF noise vs. regular noise on the CIFAR-10 dataset. In the appendix, we also include further derivations for non-isotropic SMLD models. GFFs are just one example of a well known class of models that are a subset of non-isotropic Gaussian distributions. 
%and we hope that our work will facilitate the examination of other settings.  
In the same way that other work has examined non-Gaussian distributions such as the Gamma distribution~\cite{Nachmani2022DDGM}, Poisson distribution~\cite{xu2022Poisson}, and Heat dissipation processes~\cite{Rissanen2022heat}, we hope that our work here may lay the foundation for other new denoising diffusion formulations. % Other more sophisticated methods such as neural diffusion processes~\cite{Dutordoir2022NDP}     

\newpage

\bibliographystyle{plain}
\bibliography{ref}

\newpage
\appendix

\section{Denoising Diffusion Probabilistic Models (DDPM) \cite{ho2020ddpm}}
\label{sec:ddpm}

\subsection{Forward Process}

In DDPM, for a fixed sequence of positive scales $0 < \beta_1 < \cdots < \beta_L < 1$, $\balpha_t = \prod_{s=1}^t (1 - \beta_s)$, and a noise sample $\rvepsilon \sim \gN(\vzero, \rmI)$, the cumulative ``forward'' noising process is:
\begin{align}
% q_{t} (\rvx_t \mid \rvx_{t-1}) &= \gN(\rvx_t \mid \sqrt{1 - \beta_t} \rvx_{t-1}, \beta_t \rmI)
% \implies
% \rvx_t = \sqrt{1 - \beta_t}\rvx_{t-1} + \sqrt{\beta_t} \rvz_t \\
&q_t (\rvx_t \mid \rvx_0) = \gN(\sqrt{\balpha_t} \rvx_0, (1 - \balpha_t) \rmI) \implies \rvx_t = \sqrt{\balpha_t}\rvx_0 + \sqrt{1 - \balpha_t} \rvepsilon
\label{eq:DDPM_noise}
% \\
% \implies &\log q (\rvx_t \mid \rvx_0) = \log C - \frac{1}{2(1 - \balpha_t)}(\rvx_t - \sqrt{\balpha_t}\rvx_0)^\trns(\rvx_t - \sqrt{\balpha_t}\rvx_0)
\\
\implies \text{Score }\ &\rvs = \nabla_{\rvx_t} \log q_{\balpha_t} (\rvx_t \mid \rvx_0) = -\frac{1}{(1 - \balpha_t)}(\rvx_t - \sqrt{\balpha_t} \rvx_0) = -\frac{1}{\sqrt{1 - \balpha_t}}\rvepsilon
\label{eq:Score_DDPM}
\end{align}
The noise $\rvepsilon$ is estimated using a neural network $\rvepsilon_\rvtheta(\rvx_t, t)$. Thus,
\begin{gather*}
\hat\rvx_0 = \frac{1}{\sqrt{\balpha_t}}(\rvx_t - \sqrt{1 - \balpha_t}\rvepsilon_{\rvtheta}(\rvx_t, t))
\label{eq:x0hat}
\numberthis
\\ 
[\because \rvx_t = \sqrt{\balpha_t}\rvx_0 + \sqrt{1 - \balpha_t} \rvepsilon \text{ from \cref{eq:DDPM_noise}, and loss is minimized when } \rvepsilon_{\rvtheta^*}(\rvx_t) = \rvepsilon]
\end{gather*}

\subsection{Objective function for DDPM}

The objective function for DDPM at noise level $\sigma$ is:
\begin{align}
&\loss^\DDPM(\rvtheta; \balpha_t) \triangleq\ \frac{1}{2} \EE_{q_{\balpha_t}(\rvx_t \mid \rvx) p(\rvx)} \bigg[ \Norm{\rvs_\rvtheta (\rvx_t, \balpha_t) + \frac{1}{(1 - \balpha_t)}(\rvx_t - \sqrt{\balpha_t} \rvx_0) }_2^2 \bigg]
\end{align}
The overall loss is the weighted sum of the losses at each step:
\begin{align}
& \gL(\rvtheta; \{\balpha_t\}_{t=1}^{L}) \triangleq \frac{1}{L} \sum_{t=1}^{L} \lambda(\alpha_t) \ \loss(\rvtheta; \balpha_t)
\end{align}
The weight $\lambda$ is the inverse of the variance of the score.

\subsection{Variance of score for DDPM}
\begin{align*}
&\EE\left[ \Norm{ \nabla_{\rvx_t} \log q_{t} (\rvx_t \mid \rvx_0) }_2^2 \right]
= \EE\left[ \Norm{ -\frac{\rvx_t - \sqrt{\balpha_t} \rvx_0}{(1 - \balpha_t)}}_2^2 \right]
&= \EE\left[ \Norm{ \frac{\sqrt{1 - \balpha_t}\rvepsilon}{(1 - \balpha_t)} }_2^2 \right]
= \frac{1}{1 - \balpha_t}
% \EE\left[ \Norm{\rvepsilon}_2^2 \right] 
% = \frac{1}{1 - \balpha_t} \dim(\rvepsilon)
\numberthis
\end{align*}

\subsection{Overall objective function for DDPM}

The overall objective function in \cite{ho2020ddpm} used $\lambda(\balpha_t) \propto 1/\EE\big[ \Norm{ \nabla_{\rvx_t} \log q_{t} (\rvx_t \mid \rvx_0) }_2^2 \big] = 1 - \balpha_t$:
\begin{align*}
\gL^\DDPM(\rvtheta; \{\balpha_t\}_{t=1}^{L}) &\triangleq \frac{1}{2L} \sum_{t=1}^{L} \EE_{q_{\balpha_t}(\rvx_t \mid \rvx_0) p(\rvx_0)} \bigg[ \Norm{\sqrt{1 - \balpha_t} \rvs_\rvtheta (\rvx_t, \balpha_t) + \frac{(\rvx_t - \sqrt{\balpha_t} \rvx_0)}{\sqrt{1 - \balpha_t}} }_2^2 \bigg] \\
&= \frac{1}{2L} \sum_{t=1}^{L} \EE_{q_{\balpha_t}(\rvx_t \mid \rvx_0) p(\rvx_0)} \bigg[ \Norm{\sqrt{1 - \balpha_t} \rvs_\rvtheta (\rvx_t, \balpha_t) + \rvepsilon }_2^2 \bigg]
\numberthis
\end{align*}

\subsection{Smarter DDPM score estimation}

A smarter score model recognizes that the score is a factor of $\rvepsilon$ from \cref{eq:Score_DDPM}, hence only $\rvepsilon$ needs to be estimated:
\begin{align}
\rvs_\rvtheta(\rvx_t, \balpha_t) = -\frac{1}{\sqrt{1 - \balpha_t}} \rvepsilon_\rvtheta(\rvx_t, \balpha_t)
\end{align}

In this case, the overall objective function changes to:
\begin{align*}
&\gL^\DDPM(\rvtheta; \{\balpha_t\}_{t=1}^{L}) 
\triangleq \frac{1}{2L} \sum_{t=1}^{L} \EE_{q_{\balpha_t}(\rvx_t \mid \rvx_0) p(\rvx_0)} \left[ \Norm{-\rvepsilon_\rvtheta (\rvx_t, \balpha_t) + \rvepsilon }_2^2 \right] \\
&= \frac{1}{2L} \sum_{t=1}^{L} \EE_{q_{\balpha_t}(\rvx_t \mid \rvx_0) p(\rvx_0)} \left[ \Norm{ \rvepsilon - \rvepsilon_\rvtheta (\sqrt{\balpha_t}\rvx_0 + \sqrt{1 - \balpha_t}\rvepsilon, \balpha_t)}_2^2 \right]
\numberthis
\end{align*}

This is eq 14 in the DDPM paper. The DDPM paper retains conditioning of $\rvepsilon_\rvtheta$ on $\balpha_t$, but SMLD omits it.

\subsection{Sampling in DDPM}

The reverse probability is given by:
\begin{gather*}
q(\rvx_{t-1} \mid \rvx_t, \rvx_0) = \gN(\rvx_{t-1} \mid \tilde\vmu_t(\rvx_t, \rvx_0), \tilde\beta_i\rmI) \text{  where} \\
\tilde\vmu_t(\rvx_t, \rvx_0) = \frac{\sqrt{\balpha_{t-1}}\beta_t}{1 - \balpha_t}\rvx_0 + \frac{\sqrt{1 - \beta_t}(1 - \balpha_{t-1})}{1 - \balpha_t}\rvx_t ;\quad \tilde\beta_t = \frac{1 - \balpha_{t-1}}{1 - \balpha_t}\beta_t
\numberthis
\label{eq:hatx0}
\end{gather*}
Hence, considering $\hat\rvx_0$ estimated from $\rvx_t$ using \cref{eq:x0hat}:
\begin{gather*}
\rvx_{t-1} = \frac{\sqrt{\balpha_{t-1}}\beta_t}{1 - \balpha_t}\hat\rvx_0 + \frac{\sqrt{1 - \beta_t}(1 - \balpha_{t-1})}{1 - \balpha_t}\rvx_t + \sqrt{\tilde\beta_t} \rvz_t
\numberthis
\end{gather*}

\cite{ho2020ddpm} breaks down the reversal into 2 steps:
\begin{gather*}
\hat\rvx_0 = \frac{1}{\sqrt{\balpha_t}}(\rvx_t - \sqrt{1 - \balpha_t}\rvepsilon_{\rvtheta^*}(\rvx_t)) \\
\rvx_{t-1} = \frac{\sqrt{\balpha_{t-1}}\beta_t}{1 - \balpha_t}\hat\rvx_0 + \frac{\sqrt{1 - \beta_t}(1 - \balpha_{t-1})}{1 - \balpha_t}\rvx_t + \sqrt{\tilde\beta_t} \rvz_t
\numberthis
\end{gather*}

\subsection{Sampling using DDIM}

DDIM~\cite{song2020ddim} uses this sampling:
\begin{align*}
q_{\balpha_L}(\rvx_L \mid \rvx_0) &= \gN(\rvx_L \mid \sqrt{\balpha_L}\rvx_0, (1 - \balpha_L)\rmI)
\numberthis
\\
q(\rvx_{t-1} \mid \rvx_t, \rvx_0) &= \gN\left( \rvx_{t-1} \mid \sqrt{\balpha_{t-1}}\rvx_0 + \sqrt{1 - \balpha_{t-1}}\frac{\rvx_t - \sqrt{\balpha_t}\rvx_0}{\sqrt{1 - \balpha_t}}, \vzero \right)
\numberthis
\\
\implies q_{\balpha_t}(\rvx_{t} \mid \rvx_0) &= \gN\left( \rvx_{t} \mid \sqrt{\balpha_{t}}\rvx_0, (1 - \balpha_{t})\rmI \right)
\end{align*}
% From 2.115 in \cite{bishop2006pattern}:
% \begin{align*}
% p(\rvu) &= \gN(\rvu \mid \vmu, \mLambda^{-1}),
% \\
% p(\rvv \mid \rvu) &= \gN(\rvv \mid \rmA\rvu + \rvb, \rmL^{-1})
% \\
% \implies
% p(\rvv) &= \gN(\rvv \mid \rmA\vmu + \rvb, \rmL^{-1} + \rmA \mLambda^{-1} \rmA^\trns)
% \end{align*}
% Proof by induction:
% \begin{align*}
% \text{We know that } q_{\balpha_L}(\rvx_L \mid \rvx_0) &= \gN(\rvx_L \mid \sqrt{\balpha_L}\rvx_0, (1 - \balpha_L)\rmI)
% \\
% \text{Assuming } q_{\balpha_t}(\rvx_t \mid \rvx_0) &= \gN(\rvx_t \mid \sqrt{\balpha_t}\rvx_0, (1 - \balpha_t)\rmI)
% \\
% q(\rvx_{t-1} \mid \rvx_t, \rvx_0) &= \gN\left( \rvx_{t-1} \mid \sqrt{\balpha_{t-1}}\rvx_0 + \sqrt{1 - \balpha_{t-1}}\frac{\rvx_t - \sqrt{\balpha_t}\rvx_0}{\sqrt{1 - \balpha_t}}, \vzero \right)
% \\
% \implies q_{\balpha_{t}}(\rvx_{t-1} \mid \rvx_0) &= \gN\bigg( \rvx_{t-1} \mid \sqrt{\balpha_{t-1}}\rvx_0 + \sqrt{1 - \balpha_{t-1}}\frac{\sqrt{\balpha_t}\rvx_0 - \sqrt{\balpha_t}\rvx_0}{\sqrt{1 - \balpha_t}},\\
% &\qquad\qquad \vzero + \frac{1 - \balpha_{t-1}}{1 - \balpha_{t}}(1 - \balpha_t)\rmI \bigg)\\
% &= \gN\left( \rvx_{t-1} \mid \sqrt{\balpha_{t-1}}\rvx_0, (1 - \balpha_{t-1})\rmI \right)
% \end{align*}
Hence:
\begin{align*}
\hat\rvx_0 &= \frac{1}{\sqrt{\balpha_t}}(\rvx_t - \sqrt{1 - \balpha_t}\rvepsilon_{\rvtheta^*}(\rvx_t))\\
\rvx_{t-1} &= \sqrt{\balpha_{t-1}} \hat\rvx_0 + \sqrt{1 - \balpha_{t-1}} \rvepsilon_{\rvtheta^*}(\rvx_t)
\numberthis
\end{align*}

\subsection{Expected Denoised Sample}

From \cite{saremi2019neb}, assuming isotropic Gaussian noise of variance $1-\balpha_t$, we know that the expected denoised sample $\rvx^*(\rvx_t, \balpha_t) \triangleq \EE_{\rvx \sim q_{\balpha_t}(\rvx \mid \rvx_t)}[\rvx]$ and the optimal score $\rvs_{\rvtheta^*}(\rvx_t, \balpha_t)$ are related as:
\begin{align*}
&\rvs_{\rvtheta^*}(\rvx_t, \balpha_t) = \EE\left[ \Norm{ \nabla_{\rvx_t} \log q_{\balpha_t} (\rvx_t \mid \rvx) }_2^2 \right] (\rvx^* (\rvx_t, \balpha_t) - \rvx_t) \\
&\implies \rvs_{\rvtheta^*}(\rvx_t, \balpha_t) = \frac{1}{1 - \balpha_t}(\rvx^*(\rvx_t, \balpha_t) - \rvx_t) \\
&\implies \rvx^*(\rvx_t, \balpha_t) = \rvx_t + (1 - \balpha_t)\ \rvs_{\rvtheta^*}(\rvx_t, \balpha_t) = \rvx_t - \sqrt{1 - \balpha_t} \ \rvepsilon_{\rvtheta^*}(\rvx_t)
\numberthis
\end{align*}

\subsection{SDE formulation : Variance Preserving (VP) SDE}

The above processes can be written in terms of stochastic differential equations.

Forward process:
\begin{align*}
\rvx_t &= \sqrt{1 - \beta_t}\rvx_{t-1} + \sqrt{\beta_t} \rvz_{t-1} \\
\implies \rvx(t + \Delta t) &= \sqrt{1 - \beta(t + \Delta t) \Delta t}\ \rvx(t) + \sqrt{\beta(t + \Delta t) \Delta t}\ \rvz(t) \\
&\approx \left(1 - \frac{1}{2}\beta(t + \Delta t)\Delta t\right) \rvx(t) + \sqrt{\beta(t + \Delta t) \Delta t}\ \rvz(t) \\
&\approx \rvx(t) - \frac{1}{2}\beta(t)\Delta t\ \rvx(t) + \sqrt{\beta(t) \Delta t}\ \rvz(t) \\
\implies \D \rvx &= -\frac{1}{2}\beta(t)\rvx\ \D t + \sqrt{\beta(t)}\ \D \rvw
\numberthis
\end{align*}

% \begin{align*}
% &\D \rvx = \rvf(\rvx, t) \D t + \rmL(\rvx, t) \D \rvw
% \implies \frac{\D \vmu}{\D t} = \EE_\rvx[\rvf(\rvx, t)], \\
% &\frac{\D \textbf{Cov}[\rvx]}{\D t} = \EE_\rvx[\rvf(\rvx, t) (\rvx - \vmu)^\trns] + \EE_\rvx[(\rvx - \vmu) \rvf(\rvx, t)^\trns] + \EE_\rvx[\rmL(\rvx, t) \rmQ \rmL^\trns(\rvx, t)]
% \end{align*}
% where $\rvw$ is Brownian motion, $\rmQ$ is the PSD of $\rvw$. For Gaussian noise, $\rmQ = \rmI$.

Mean (from eq. 5.50 in Sarkka \& Solin (2019)):
\begin{align*}
&\D \rvx = \rvf\ \D t + \rmG\ \D \rvw \implies \frac{\D \vmu}{\D t} = \EE_\rvx[\rvf] \\
&\therefore \frac{\D \vmu_{\DDPM}(t)}{\D t} = \EE_\rvx[-\frac{1}{2}\beta(t)\rvx] = -\frac{1}{2}\beta(t)\EE_\rvx(\rvx) = -\frac{1}{2}\beta(t)\vmu_{\DDPM}(t) \\
&\implies \frac{\D \vmu_{\DDPM}(t)}{\vmu_{\DDPM}(t)} = -\frac{1}{2}\beta(t) \D t \implies \log \vmu_{\DDPM}(t) \vert_{0}^{t} = -\frac{1}{2}\int_0^t \beta(s) \D s \\
&\implies \log \vmu_{\DDPM}(t) - \log \vmu(0) = -\frac{1}{2}\int_0^t \beta(s) \D s 
\implies \log \frac{\vmu_{\DDPM}(t)}{\vmu(0)} = -\frac{1}{2}\int_0^t \beta(s) \D s \\
&\implies \vmu_{\DDPM}(t) = \vmu(0)\ e^{-\frac{1}{2}\int_0^t \beta(s) \D s} \\
\end{align*}

Covariance (from eq. 5.51 in Sarkka \& Solin (2019)):
\begin{align*}
&\D \rvx = \rvf\ \D t + \rmG\ \D \rvw \implies \frac{\D \mSigma_{\text{cov}}}{\D t} = \EE_\rvx[\rvf (\rvx - \vmu)^\trns] + \EE_\rvx[(\rvx - \vmu) \rvf^\trns] + \EE_\rvx[\rmG\rmG^\trns] \\
&\therefore \frac{\D \mSigma_{\DDPM}(t)}{\D t} = \EE_\rvx[-\frac{1}{2}\beta(t)\rvx\rvx^\trns] + \EE_\rvx[\rvx(-\frac{1}{2}\beta(t)\rvx)^\trns] + \EE_\rvx[\sqrt{\beta(t)}\rmI\sqrt{\beta(t)}\rmI] \\
&\qquad\qquad\qquad = -\beta(t)\mSigma_{\DDPM}(t) + \beta(t)\rmI = \beta(t)(\rmI - \mSigma_{\DDPM}(t)) \\
&\implies \frac{\D \mSigma_{\DDPM}(t)}{\rmI - \mSigma_{\DDPM}(t)} = \beta(t) \D t \implies -\log(\rmI - \mSigma_{\DDPM}(t)) \vert_{0}^{t} = \int_0^t \beta(s) \D s \\
&\implies -\log(\rmI - \mSigma_{\DDPM}(t)) + \log(\rmI - \mSigma_{\rvx}(0)) = \int_0^t \beta(s) \D s \\
&\implies \frac{\rmI - \mSigma_{\DDPM}(t)}{\rmI - \mSigma_{\rvx}(0)} = e^{-\int_0^t \beta(s) \D s} \implies \mSigma_{\DDPM}(t) = \rmI - e^{-\int_0^t \beta(s) \D s} (\rmI - \mSigma_{\rvx}(0)) \\
&\implies \mSigma_{\DDPM}(t) = \rmI + e^{-\int_0^t \beta(s) \D s} (\mSigma_{\rvx}(0) - \rmI)
\end{align*}

For each data point $\rvx(0)$, $\vmu(0) = \rvx(0)$, $\mSigma_{\rvx}(0) = \vzero$:
\begin{align*}
&\implies \vmu_{\DDPM}(t) = \rvx(0)\ e^{-\frac{1}{2}\int_0^t \beta(s) \D s}, \\
&\qquad \quad \mSigma_{\DDPM}(t) = \rmI + e^{-\int_0^t \beta(s) \D s} (\vzero - \rmI) = \rmI - \rmI e^{-\int_0^t \beta(s) \D s} \\
&\therefore p_{0t}(\rvx(t) \mid \rvx(0)) = \gN \left(\rvx(t) \mid \rvx(0)\ e^{-\frac{1}{2}\int_0^t \beta(s) \D s}, \rmI - \rmI e^{-\int_0^t \beta(s) \D s} \right)
\end{align*}

Beta schedule, linear:
\begin{align*}
&\beta(t) = \beta_{\min} + t (\beta_{\max} - \beta_{\min}) \implies \int_0^t \beta(s) \D s = t \beta_{\min} + \frac{t^2}{2}(\beta_{\max} - \beta_{\min})
\end{align*}

\newpage

\section{Non-isotropic DDPM (NI-DDPM)}
\label{sec:niddpm}

\subsection{Score for NI-DDPM}
\label{sec:niddpm_score}

For a fixed sequence of positive scales $0 < \beta_1 < \cdots < \beta_L < 1$, $\balpha_t = \prod_{s=1}^i (1 - \beta_s)$,
\begin{align}
&q_{\balpha_t}^{\GFFDDPM} (\rvx_t \mid \rvx_{t-1}) = \gN(\rvx_t \mid \sqrt{1 - \beta_t} \rvx_{t-1}, \beta_t \mSigma) \\
&\implies \rvx_t = \sqrt{1 - \beta_t}\rvx_{t-1} + \sqrt{\beta_t} \sqrt{\mSigma} \rvz_{t-1} \\
&q_{\balpha_t}^{\GFFDDPM} (\rvx_t \mid \rvx_0) = \gN(\rvx_t \mid \sqrt{\balpha_t} \rvx_0, (1 - \balpha_t) \mSigma) \\
&\implies \rvx_t = \sqrt{\balpha_t}\rvx_0 + \sqrt{1 - \balpha_t} \sqrt{\mSigma} \rvepsilon
\implies \rvepsilon = \sqrt{\mSigma^{-1}}\frac{\rvx_t - \sqrt{\balpha_t}\rvx_0}{\sqrt{1 - \balpha_t}}
\label{eq:GFF_DDPM_noise}
\\
&\implies \nabla_{\rvx_t} \log q_{\balpha_t}^{\GFFDDPM} (\rvx_t \mid \rvx_0) = -\mSigma^{-1}\frac{\rvx_t - \sqrt{\balpha_t} \rvx_0}{1 - \balpha_t} = -\frac{1}{\sqrt{1 - \balpha_t}}\sqrt{\mSigma^{-1}}\rvepsilon
\label{eq:score-noise-NIDDPM}
\end{align}

Derivation of the score value:
\begin{align*}
&q_{\balpha_t}^{\GFFDDPM} (\rvx_t \mid \rvx_0) = \gN(\rvx_t \mid \sqrt{\balpha_t} \rvx_0, (1 - \balpha_t) \mSigma) \\
&= \frac{1}{(2\pi)^{D/2}((1 - \balpha_t)|\mSigma|)^{1/2}}\exp\left( -\frac{1}{2(1 - \balpha_t)}(\rvx_t - \sqrt{\balpha_t}\rvx_0)^\trns\mSigma^{-1}(\rvx_t - \sqrt{\balpha_t}\rvx_0) \right) \\
\implies &\log q_{\balpha_t}^{\GFFDDPM} (\rvx_t \mid \rvx_0) = -\log ((2\pi)^{D/2}((1 - \balpha_t)|\mSigma|)^{1/2})
\\
&\qquad\qquad\qquad\qquad\qquad\quad - \frac{1}{2(1 - \balpha_t)}(\rvx_t - \sqrt{\balpha_t}\rvx_0)^\trns\mSigma^{-1}(\rvx_t - \sqrt{\balpha_t}\rvx_0) \\
\implies &\nabla_{\rvx_t} \log q_{\balpha_t}^{\GFFDDPM} (\rvx_t \mid \rvx_0) = -\frac{1}{\cancel{2}(1 - \balpha_t)}\cancel{2}\mSigma^{-1}(\rvx_t - \sqrt{\balpha_t}\rvx_0) = -\frac{1}{\sqrt{1 - \balpha_t}}\sqrt{\mSigma^{-1}}\rvepsilon
\end{align*}

\subsection{Objective function for NI-DDPM}

The objective function for score estimation in NI-DDPM at noise level $\balpha_t$ is:
\begin{align*}
\loss^{\GFFDDPM}(\rvtheta; \balpha_t) &\triangleq\ \frac{1}{2} \EE_{q_{\balpha_t}(\rvx_t \mid \rvx_0) p(\rvx_0)} \left[ \Norm{\rvs_\rvtheta (\rvx_t, \balpha_t) + \mSigma^{-1} \frac{\rvx_t - \sqrt{\balpha_t} \rvx_0}{1 - \balpha_t} }_2^2 \right]
\numberthis \\
&\triangleq\ \frac{1}{2} \EE_{q_{\balpha_t}(\rvx_t \mid \rvx_0) p(\rvx_0)} \left[ \Norm{\rvs_\rvtheta (\rvx_t, \balpha_t) +  \frac{1}{\sqrt{1 - \balpha_t}}\sqrt{\mSigma^{-1}}\rvepsilon }_2^2 \right]
\end{align*}

\subsection{Expected value of score for NI-DDPM}
\begin{align*}
&\EE\left[ \Norm{ \nabla_{\rvx_t} \log q_{\balpha_t}^{\GFFDDPM} (\rvx_t \mid \rvx_0) }_2^2 \right]
= \EE\left[ \Norm{ -\mSigma^{-1}\frac{\rvx_t - \sqrt{\balpha_t} \rvx_0}{1 - \balpha_t} }_2^2 \right] \\
&= \EE\left[ \Norm{ \mSigma^{-1}\frac{\sqrt{1 - \balpha_t}\sqrt{\mSigma}\rvepsilon}{1 - \balpha_t} }_2^2 \right]
= \frac{1}{1 - \balpha_t}\mSigma^{-1}\EE\left[ \Norm{\rvepsilon}_2^2 \right] = \frac{1}{1 - \balpha_t} \mSigma^{-1}
\numberthis
\end{align*}

\subsection{Overall objective function for NI-DDPM}
\label{subsec:niddpm_objective}

\begin{align*}
& \gL(\rvtheta; \{\balpha_t\}_{t=1}^{L}) \triangleq \frac{1}{L} \sum_{t=1}^{L} \lambda(\alpha_t) \ \loss^{\GFFDDPM}(\rvtheta; \balpha_t)
\end{align*}

$\lambda$ inverse of the variance i.e. $ \lambda(\balpha_t) = (1 - \balpha_t)\mSigma \implies$
\unboldmath
\begin{align*}
\gL_{\GFFDDPM}(\rvtheta; \{\balpha_t\}_{t=1}^{L})
\triangleq &\frac{1}{2L} \sum_{t=1}^{L} \EE_{q_{\balpha_t}(\rvx_t \mid \rvx_0) p(\rvx_0)} \left[ \Norm{ \sqrt{1 - \balpha_t}\sqrt{\mSigma} \rvs_\rvtheta (\rvx_t, \balpha_t) + \sqrt{\mSigma^{-1}}\frac{(\rvx_t - \sqrt{\balpha_t} \rvx_0)}{\sqrt{1 - \balpha_t}} }_2^2 \right] \\
&= \frac{1}{2L} \sum_{t=1}^{L} \EE_{q_{\balpha_t}(\rvx_t \mid \rvx_0) p(\rvx_0)} \left[ \Norm{ \sqrt{1 - \balpha_t}\sqrt{\mSigma} \rvs_\rvtheta (\rvx_t, \balpha_t) + \rvepsilon }_2^2 \right]
\numberthis
\end{align*}

% (b) \boldmath
% $ \lambda_b(\balpha_t) = (1 - \balpha_t)$
% \unboldmath
% \begin{align*}
% &\gL_{\GFFDDPM}^b(\rvtheta; \{\balpha_t\}_{t=1}^{L})
% \triangleq \frac{1}{2L} \sum_{t=1}^{L} \EE_{q_{\balpha_t}(\rvx_t \mid \rvx_0) p(\rvx_0)} \left[ \Norm{ \sqrt{1 - \balpha_t} \rvs_\rvtheta (\rvx_t, \balpha_t) + \mSigma^{-1}\frac{(\rvx_t - \sqrt{\balpha_t} \rvx_0)}{\sqrt{1 - \balpha_t}} }_2^2 \right] \\
% &= \frac{1}{2L} \sum_{t=1}^{L} \EE_{q_{\balpha_t}(\rvx_t \mid \rvx_0) p(\rvx_0)} \left[ \Norm{ \sqrt{1 - \balpha_t} \rvs_\rvtheta (\rvx_t, \balpha_t) + \sqrt{\mSigma^{-1}} \rvepsilon }_2^2 \right]
% \numberthis
% \end{align*}

% (c) \boldmath
% $ \lambda_c(\balpha_t) = (1 - \balpha_t)\mSigma^2$
% \unboldmath
% \begin{align*}
% &\gL_{\GFFDDPM}^c(\rvtheta; \{\balpha_t\}_{t=1}^{L})
% \triangleq \frac{1}{2L} \sum_{t=1}^{L} \EE_{q_{\balpha_t}(\rvx_t \mid \rvx_0) p(\rvx_0)} \left[ \Norm{ \sqrt{1 - \balpha_t} \mSigma \rvs_\rvtheta (\rvx_t, \balpha_t) + \frac{(\rvx_t - \sqrt{\balpha_t} \rvx_0)}{\sqrt{1 - \balpha_t}} }_2^2 \right] \\
% &= \frac{1}{2L} \sum_{t=1}^{L} \EE_{q_{\balpha_t}(\rvx_t \mid \rvx_0) p(\rvx_0)} \left[ \Norm{ \sqrt{1 - \balpha_t} \mSigma \rvs_\rvtheta (\rvx_t, \balpha_t) + \sqrt{\mSigma} \rvepsilon }_2^2 \right]
% \numberthis
% \end{align*}

\subsection{NI-DDPM score estimation using noise estimation}

\textbf{A score model} that matches the actual score-noise relationship in \cref{eq:score-noise-NIDDPM} is:
\begin{align}
\rvs_\rvtheta(\rvx_t, \balpha_t) = -\sqrt{\mSigma^{-1}}\frac{\rvepsilon_\rvtheta^{(1)}(\rvx_t)}{\sqrt{1 - \balpha_t}}
\end{align}

In this case, the overall objective function changes to:
\begin{align*}
&\gL_{\GFFDDPM}(\rvtheta; \{\balpha_t\}_{t=1}^{L}) \triangleq \frac{1}{2L} \sum_{t=1}^{L} \EE_{q_{\balpha_t}(\rvx_t \mid \rvx_0) p(\rvx_0)} \bigg[ \Norm{\rvepsilon - \rvepsilon_\rvtheta^{(1)} (\rvx_t)}_2^2 \bigg]
\numberthis 
% \\
% &\gL_{\GFFDDPM}^{(1)b}(\rvtheta; \{\balpha_t\}_{t=1}^{L}) \triangleq \frac{1}{2L} \sum_{t=1}^{L} \EE_{q_{\balpha_t}(\rvx_t \mid \rvx_0) p(\rvx_0)} \bigg[ \Norm{\sqrt{\mSigma^{-1}}\rvepsilon  - \sqrt{\mSigma^{-1}}\rvepsilon_\rvtheta^{(1)} (\rvx_t)}_2^2 \bigg]
% \numberthis \\
% &\gL_{\GFFDDPM}^{(1)c}(\rvtheta; \{\balpha_t\}_{t=1}^{L}) \triangleq \frac{1}{2L} \sum_{t=1}^{L} \EE_{q_{\balpha_t}(\rvx_t \mid \rvx_0) p(\rvx_0)} \bigg[ \Norm{\sqrt{\mSigma}\rvepsilon - \sqrt{\mSigma}\rvepsilon_\rvtheta^{(1)} (\rvx_t) }_2^2 \bigg]
% \numberthis
\end{align*}

\subsection{Sampling in NI-DDPM}
\label{subsec:sampling_niddpm}

% \cite{ho2020ddpm} breaks down the reversal into 2 steps:
% \begin{align*}
% q_{t}^{\GFFDDPM} (\rvx_t \mid \rvx_0) &= \gN(\rvx_t \mid \sqrt{\balpha_t} \rvx_0, (1 - \balpha_t) \mSigma)
% % \\
% % \implies \hat\rvx_0 &= \frac{1}{\sqrt{\balpha_t}}(\rvx_t - \sqrt{1 - \balpha_t}\sqrt{\mSigma}\rvepsilon)
% \\
% \implies \hat\rvx_0 &= \frac{1}{\sqrt{\balpha_t}}(\rvx_t - \sqrt{1 - \balpha_t}\sqrt{\mSigma}\rvepsilon_{\rvtheta^*}(\rvx_t))
% \numberthis
% \\
% \implies \hat\rvx_0^{(2)} &= \frac{1}{\sqrt{\balpha_t}}(\rvx_t - \sqrt{1 - \balpha_t}\mSigma\rvepsilon_{\rvtheta^*}^{(2)}(\rvx_t))
% \numberthis \\
% \implies \hat\rvx_0^{(3)} &= \frac{1}{\sqrt{\balpha_t}}(\rvx_t - \sqrt{1 - \balpha_t}\rvepsilon_{\rvtheta^*}^{(3)}(\rvx_t))
% \numberthis
% \end{align*}
We compute the parameters of the reverse process using Bishop (2006) 2.116, by additionally conditioning on $\rvx_0$:
\begin{subequations}
\begin{empheq}[box=\widefbox]{align*}
p(\rvu) &= \gN(\rvu \mid \vmu, \mLambda^{-1}),
\\
p(\rvv \mid \rvu) &= \gN(\rvv \mid \rmA\rvu + \rvb, \rmL^{-1})
\\
\implies
p(\rvv) &= \gN(\rvv \mid \rmA\vmu + \rvb, \rmL^{-1} + \rmA \mLambda^{-1} \rmA^\trns),
\\
\implies p(\rvu \mid \rvv) &= \gN(\rvu \mid \rmC(\rmA^\trns \rmL (\rvv - \rvb) + \mLambda \vmu), \rmC), \rmC = (\mLambda + \rmA^\trns \rmL \rmA)^{-1}
\end{empheq}
\end{subequations}
Here, $\rvu = \rvx_{t-1} | \rvx_0, \rvv = \rvx_{t}$:
\begin{align*}
q_{\balpha_{t-1}}(\rvx_{t-1} \mid \rvx_{0}) &= \gN(\rvx_{t-1} \mid \sqrt{\balpha_{t-1}}\rvx_0, (1 - \balpha_{t-1})\mSigma)
\\
q_{\balpha_t}(\rvx_t \mid \rvx_{t-1}, \rvx_0) &= \gN(\rvx_t \mid \sqrt{1 - \beta_t}\rvx_{t-1}, \beta_t\mSigma)
\\
\implies \rvx &= \rvx_{t-1} \mid \rvx_0, \rvy = \rvx_t \mid \rvx_0, \text{ need } p(\rvx \mid \rvy) = q(\rvx_{t-1} \mid \rvx_t, \rvx_0)
\\
\implies \vmu &= \sqrt{\balpha_{t-1}}\rvx_0,  \mLambda^{-1} = (1 - \balpha_{t-1})\mSigma, \rmA = \sqrt{1 - \beta_t}, \rvb = \vzero, \rmL^{-1} = \beta_t\mSigma
\\
\implies \rmC &= \left(\frac{1}{1 - \balpha_{t-1}}\mSigma^{-1} + (1 - \beta_t)\frac{1}{\beta_t}\mSigma^{-1}\right)^{-1} = \left(\frac{\cancel{\beta_t} + 1 - \cancel{\beta_{t}} - \alpha_t}{(1 - \balpha_{t-1})\beta_t}\mSigma^{-1}\right)^{-1} \\
&= \frac{1 - \balpha_{t-1}}{1 - \balpha_t}\beta_t\mSigma = \tilde\beta_t\mSigma
\\
\implies \rmC(\rmA^\trns \rmL (\rvy - \rvb) &+ \mLambda \vmu) = \frac{1 - \balpha_{t-1}}{1 - \balpha_t}\beta_t\mSigma \left(\sqrt{1 - \beta_t}\frac{1}{\beta_t}\mSigma^{-1}\rvx_{t} + \frac{1}{1 - \balpha_{t-1}}\mSigma^{-1}\sqrt{\balpha_{t-1}}\rvx_0\right)\\
&= \frac{\sqrt{\balpha_{t-1}}\beta_t}{1 - \balpha_t}\rvx_0 + \frac{\sqrt{1 - \beta_t}(1 - \balpha_{t-1})}{1 - \balpha_t}\rvx_t
\end{align*}

Thus, the parameters of the distribution of the reverse process are:
\begin{align*}
\therefore &q(\rvx_{t-1} \mid \rvx_t, \rvx_0) = \gN(\rvx_{t-1} \mid \tilde\vmu_t(\rvx_t, \rvx_0), \tilde\beta_t\mSigma) \text{ , where} \\
&\tilde\vmu_t(\rvx_t, \rvx_0) = \frac{\sqrt{\balpha_{t-1}}\beta_t}{1 - \balpha_t}\rvx_0 + \frac{\sqrt{1 - \beta_t}(1 - \balpha_{t-1})}{1 - \balpha_t}\rvx_t ;\quad \tilde\beta_t = \frac{1 - \balpha_{t-1}}{1 - \balpha_t}\beta_t
\numberthis\\
&\implies \rvx_{t-1}
= \tilde\vmu_t(\rvx_t, \hat\rvx_0) + \sqrt{\tilde\beta_t} \sqrt{\mSigma} \rvepsilon_t \text{ , where} \\
&\tilde\vmu_t(\rvx_t, \hat\rvx_0^{(1)}) = \frac{\sqrt{\balpha_{t-1}}\beta_t}{1 - \balpha_t}\left(\frac{1}{\sqrt{\balpha_t}}\big(\rvx_t - \sqrt{1 - \balpha_t}\sqrt{\mSigma}\rvepsilon^*\big)\right) + \frac{\sqrt{1 - \beta_t}(1 - \balpha_{t-1})}{1 - \balpha_t}\rvx_t \\
&\quad = \frac{\sqrt{\balpha_{t-1}}}{\sqrt{\balpha_t}} \frac{\beta_t}{1 - \balpha_t}\rvx_t - \frac{\sqrt{\balpha_{t-1}}}{\sqrt{\balpha_t}} \frac{\beta_t}{\sqrt{1 - \balpha_t}} \sqrt{\mSigma}\rvepsilon^* + \frac{\sqrt{1 - \beta_t}}{1 - \balpha_t}\bigg(1 - \balpha_{t-1}\bigg)\rvx_t \\
&\quad = \frac{1}{\sqrt{1 - \beta_t}} \frac{\beta_t}{1 - \balpha_t}\rvx_t + \frac{\sqrt{1 - \beta_t}}{1 - \balpha_t}\left(1 - \frac{\balpha_t}{1 - \beta_t}\right)\rvx_t - \frac{1}{\sqrt{1 - \beta_t}} \frac{\beta_t}{\sqrt{1 - \balpha_t}} \sqrt{\mSigma}\rvepsilon^* \\
&\quad = \frac{1}{\sqrt{1 - \beta_t}}\left( \frac{\beta_t}{1 - \balpha_t}\rvx_t + \frac{1 - \beta_t}{1 - \balpha_t}\left(1 - \frac{\balpha_t}{1 - \beta_t}\right)\rvx_t - \frac{\beta_t}{\sqrt{1 - \balpha_t}} \sqrt{\mSigma}\rvepsilon^* \right)\\
&\quad = \frac{1}{\sqrt{1 - \beta_t}}\left( \frac{\cancel{\beta_t} + \bcancel{1 \cancel{- \beta_t} - \balpha_t}}{\bcancel{1 - \balpha_t}}\rvx_t - \frac{\beta_t}{\sqrt{1 - \balpha_t}} \sqrt{\mSigma}\rvepsilon^* \right)\\
&\implies \tilde\vmu_t(\rvx_t, \hat\rvx_0) = \frac{1}{\sqrt{1 - \beta_t}}\left( \rvx_t - \frac{\beta_t}{\sqrt{1 - \balpha_t}} \sqrt{\mSigma}\rvepsilon_{\rvtheta^*}^{(1)}(\rvx_t) \right),
\numberthis \\
% &\qquad\quad \tilde\vmu_t(\rvx_t, \hat\rvx_0^{(2)}) = \frac{1}{\sqrt{1 - \beta_t}}\left( \rvx_t - \frac{\beta_t}{\sqrt{1 - \balpha_t}} \mSigma\rvepsilon_{\rvtheta^*}^{(2)}(\rvx_t) \right)
% \numberthis \\
% &\qquad\quad \tilde\vmu_t(\rvx_t, \hat\rvx_0^{(3)}) = \frac{1}{\sqrt{1 - \beta_t}}\left( \rvx_t - \frac{\beta_t}{\sqrt{1 - \balpha_t}} \rvepsilon_{\rvtheta^*}^{(3)}(\rvx_t) \right)
% \numberthis 
% \\
\implies \rvx_{t-1}
&= \frac{1}{\sqrt{1 - \beta_t}}\left(\rvx_t - \frac{\beta_t}{\sqrt{1 - \balpha_t}}\sqrt{\mSigma}\ \rvepsilon_{\rvtheta^*}(\rvx_t)\right) + \sqrt{\tilde\beta_t} \sqrt{\mSigma}\ \rvepsilon_t
\numberthis
\\
&= \frac{1}{\sqrt{1 - \beta_t}}\big(\rvx_t + \beta_t \mSigma\ \rvs_{\rvtheta^*}(\rvx_t, \balpha_t)\big) + \sqrt{\tilde\beta_t} \sqrt{\mSigma}\ \rvepsilon_t,
\numberthis
% \\
% \rvx_{t-1}^{(2)}
% &= \frac{1}{\sqrt{1 - \beta_t}}\left(\rvx_t - \frac{\beta_t}{\sqrt{1 - \balpha_t}}\mSigma\ \rvepsilon_{\rvtheta^*}^{(2)}(\rvx_t)\right) + \sqrt{\tilde\beta_t} \sqrt{\mSigma}\ \rvepsilon_t
% \numberthis
% \\
% &= \frac{1}{\sqrt{1 - \beta_t}}\big(\rvx_t + \beta_t \mSigma\ \rvs_{\rvtheta^*}^{(2)}(\rvx_t, \balpha_i)\big) + \sqrt{\tilde\beta_t} \sqrt{\mSigma}\ \rvepsilon_t
% \numberthis
% \\
% \rvx_{t-1}^{(3)}
% &= \frac{1}{\sqrt{1 - \beta_t}}\left(\rvx_t - \frac{\beta_t}{\sqrt{1 - \balpha_t}} \rvepsilon_{\rvtheta^*}^{(3)}(\rvx_t)\right) + \sqrt{\tilde\beta_t} \sqrt{\mSigma}\ \rvepsilon_t
% \numberthis
% \\
% &= \frac{1}{\sqrt{1 - \beta_t}}\big(\rvx_t + \beta_t \mSigma \rvs_{\rvtheta^*}^{(3)}(\rvx_t, \balpha_t)\big) + \sqrt{\tilde\beta_t} \sqrt{\mSigma}\ \rvepsilon_t
% \numberthis
\end{align*}

\subsection{Sampling using DDIM}

\begin{align*}
q_{\balpha_L}(\rvx_L \mid \rvx_0) &= \gN(\rvx_L \mid \sqrt{\balpha_L}\rvx_0, (1 - \balpha_L)\mSigma)
\numberthis
\\
q(\rvx_{t-1} \mid \rvx_t, \rvx_0) &= \gN\left( \rvx_{t-1} \mid \sqrt{\balpha_{t-1}}\rvx_0 + \sqrt{1 - \balpha_{t-1}}\frac{\rvx_t - \sqrt{\balpha_t}\rvx_0}{\sqrt{1 - \balpha_t}}, \vzero \right)
\numberthis
\\
\implies q_{\balpha_t}(\rvx_{t} \mid \rvx_0) &= \gN\left( \rvx_{t} \mid \sqrt{\balpha_{t}}\rvx_0, (1 - \balpha_{t})\mSigma \right)
\end{align*}
% From Bishop (2006) 2.115:
% \begin{align*}
% p(\rvx) &= \gN(\rvx \mid \vmu, \mLambda^{-1}),
% \\
% p(\rvy \mid \rvx) &= \gN(\rvy \mid \rmA\rvx + \rvb, \rmL^{-1})
% \\
% \implies
% p(\rvy) &= \gN(\rvy \mid \rmA\vmu + \rvb, \rmL^{-1} + \rmA \mLambda^{-1} \rmA^\trns)
% \end{align*}
% Proof by induction:
% \begin{align*}
% \text{We know that } q_{\balpha_L}(\rvx_L \mid \rvx_0) &= \gN(\rvx_L \mid \sqrt{\balpha_L}\rvx_0, (1 - \balpha_L)\mSigma)
% \\
% \text{Assuming } q_{\balpha_i}(\rvx_i \mid \rvx_0) &= \gN(\rvx_i \mid \sqrt{\balpha_i}\rvx_0, (1 - \balpha_i)\mSigma)
% \\
% q(\rvx_{i-1} \mid \rvx_i, \rvx_0) &= \gN\left( \rvx_{i-1} \mid \sqrt{\balpha_{i-1}}\rvx_0 + \sqrt{1 - \balpha_{i-1}}\frac{\rvx_i - \sqrt{\balpha_i}\rvx_0}{\sqrt{1 - \balpha_i}}, \vzero \right)
% \\
% \implies \vmu = \sqrt{\balpha_i}\rvx_0, \mLambda^{-1} &= (1 - \balpha_i)\mSigma, \rmA = \frac{\sqrt{1 - \balpha_{i-1}}}{\sqrt{1 - \balpha_i}}, \rmL^{-1} = \vzero
% \\
% \implies q_{\balpha_{i}}(\rvx_{i-1} \mid \rvx_0) &= \gN\bigg( \rvx_{i-1} \mid \sqrt{\balpha_{i-1}}\rvx_0 + \sqrt{1 - \balpha_{i-1}}\frac{\sqrt{\balpha_i}\rvx_0 - \sqrt{\balpha_i}\rvx_0}{\sqrt{1 - \balpha_i}},\\
% &\qquad\qquad \vzero + \frac{1 - \balpha_{i-1}}{\cancel{1 - \balpha_{i}}}\cancel{(1 - \balpha_i)}\mSigma \bigg)\\
% &= \gN\left( \rvx_{i-1} \mid \sqrt{\balpha_{i-1}}\rvx_0, (1 - \balpha_{i-1})\mSigma \right)
% \end{align*}
% Hence :
\begin{align*}
\implies \hat\rvx_0 &= \frac{1}{\sqrt{\balpha_t}}(\rvx_t - \sqrt{1 - \balpha_t}\sqrt{\mSigma}\rvepsilon_{\rvtheta^*}(\rvx_t))
\\
\rvx_{t-1} &= \sqrt{\balpha_{t-1}} \hat\rvx_0 + \sqrt{1 - \balpha_{t-1}} \sqrt{\mSigma} \rvepsilon_{\rvtheta^*}(\rvx_t)
\numberthis
% \\
% \implies \hat\rvx_0^{(2)} &= \frac{1}{\sqrt{\balpha_t}}(\rvx_t - \sqrt{1 - \balpha_t}\mSigma\rvepsilon_{\rvtheta^*}^{(2)}(\rvx_t))
% \\
% \rvx_{t-1}^{(2)} &= \sqrt{\balpha_{t-1}} \hat\rvx_0^{(2)} + \sqrt{1 - \balpha_{t-1}} \mSigma \rvepsilon_{\rvtheta^*}^{(2)}(\rvx_t)
% \numberthis
% \\
% \implies \hat\rvx_0^{(3)} &= \frac{1}{\sqrt{\balpha_t}}(\rvx_t - \sqrt{1 - \balpha_t}\rvepsilon_{\rvtheta^*}^{(3)}(\rvx_t))
% \\
% \rvx_{t-1}^{(3)} &= \sqrt{\balpha_{t-1}} \hat\rvx_0^{(3)} + \sqrt{1 - \balpha_{t-1}} \rvepsilon_{\rvtheta^*}^{(3)}(\rvx_t)
% \numberthis
\end{align*}

\subsection{Expected Denoised Sample}

From \cite{saremi2019neb}, assuming isotropic Gaussian noise of covariance $(1 - \balpha_t)\mSigma$, we know that the expected denoised sample $\rvx^*(\rvx_t, \balpha_t) \triangleq \EE_{\rvx \sim q_{\balpha_t}(\rvx \mid \rvx_t)}[\rvx]$ and the optimal score $\rvs_{\rvtheta^*}(\rvx_t, \balpha_t)$ are related as:
\begin{align*}
&\rvs_{\rvtheta^*}(\rvx_t, \balpha_t) = \frac{1}{1 - \balpha_t}\mSigma^{-1}(\rvx^*(\rvx_t, \balpha_t) - \rvx_t) \\
&\implies \rvx^*(\rvx_t, \balpha_t) = \rvx_t + (1 - \balpha_t) \mSigma\ \rvs_{\rvtheta^*}(\rvx_t, \balpha_t) = \rvx_t - \sqrt{1 - \balpha_t} \sqrt{\mSigma}\ \rvepsilon_{\rvtheta^*}(\rvx_t)
\numberthis 
% \\
% &\implies \rvx^{*(2)}(\rvx_t, \balpha_t) = \rvx_t + (1 - \balpha_t) \mSigma\ \rvs_{\rvtheta^*}^{(2)}(\rvx_t, \balpha_t) = \rvx_t - \sqrt{1 - \balpha_t} \mSigma\ \rvepsilon_{\rvtheta^*}^{(2)}(\rvx_t)
% \numberthis \\
% &\implies \rvx^{*(3)}(\rvx_t, \balpha_t) = \rvx_t + (1 - \balpha_t) \mSigma\ \rvs_{\rvtheta^*}^{(3)}(\rvx_t, \balpha_t) = \rvx_t - \sqrt{1 - \balpha_t} \rvepsilon_{\rvtheta^*}^{(3)}(\rvx_t)
% \numberthis
\end{align*}

\subsection{SDE formulation : Non-Isotropic Variance Preserving (NIVP) SDE}
\label{subsec:niddpm_sde}

Forward process:
\begin{align*}
\rvx_t &= \sqrt{1 - \beta_t}\rvx_{t-1} + \sqrt{\beta_t} \sqrt{\mSigma} \rvepsilon_{t-1} \\
\implies \rvx(t + \Delta t) &= \sqrt{1 - \beta(t + \Delta t) \Delta t}\ \rvx(t) + \sqrt{\beta(t + \Delta t) \Delta t}\ \sqrt{\mSigma}\rvepsilon(t) \\
&\approx \left(1 - \frac{1}{2}\beta(t + \Delta t)\Delta t\right) \rvx(t) + \sqrt{\beta(t + \Delta t) \Delta t}\ \sqrt{\mSigma}\rvepsilon(t) \\
&\approx \rvx(t) - \frac{1}{2}\beta(t)\Delta t\ \rvx(t) + \sqrt{\beta(t) \Delta t}\ \sqrt{\mSigma}\rvepsilon(t) \\
\implies \D \rvx &= -\frac{1}{2}\beta(t)\rvx\ \D t + \sqrt{\beta(t)}\sqrt{\mSigma}\ \D \rvw
\numberthis
\end{align*}

% \begin{align*}
% &\D \rvx = \rvf(\rvx, t) \D t + \rmL(\rvx, t) \D \rvw
% \implies \frac{\D \vmu}{\D t} = \EE_\rvx[\rvf(\rvx, t)], \\
% &\frac{\D \textbf{Cov}[\rvx]}{\D t} = \EE_\rvx[\rvf(\rvx, t) (\rvx - \vmu)^\trns] + \EE_\rvx[(\rvx - \vmu) \rvf(\rvx, t)^\trns] + \EE_\rvx[\rmL(\rvx, t) \rmQ \rmL^\trns(\rvx, t)]
% \end{align*}
% where $\rvw$ is Brownian motion, $\rmQ$ is the PSD of $\rvw$. For GFF noise, $\rmQ = \mSigma$.

Mean (from eq. 5.50 in Sarkka \& Solin (2019)):
\begin{align*}
&\D \rvx = \rvf\ \D t + \rmG\ \D \rvw \implies \frac{\D \vmu}{\D t} = \EE_\rvx[\rvf] \\
&\therefore \frac{\D \vmu_{\GFFDDPM}(t)}{\D t} = \EE_\rvx[-\frac{1}{2}\beta(t)\rvx] = -\frac{1}{2}\beta(t)\EE_\rvx(\rvx) = -\frac{1}{2}\beta(t)\vmu_{\GFFDDPM}(t) \\
&\implies \frac{\D \vmu_{\GFFDDPM}(t)}{\vmu_{\GFFDDPM}(t)} = -\frac{1}{2}\beta(t) \D t \implies \log \vmu_{\GFFDDPM}(t) \vert_{0}^{t} = -\frac{1}{2}\int_0^t \beta(s) \D s \\
&\implies \log \vmu_{\GFFDDPM}(t) - \log \vmu(0) = -\frac{1}{2}\int_0^t \beta(s) \D s 
\implies \log \frac{\vmu_{\GFFDDPM}(t)}{\vmu(0)} = -\frac{1}{2}\int_0^t \beta(s) \D s \\
&\implies \vmu_{\GFFDDPM}(t) = \vmu(0)\ e^{-\frac{1}{2}\int_0^t \beta(s) \D s} \\
\end{align*}

Covariance (from eq. 5.51 in Sarkka \& Solin (2019)):
\begin{align*}
&\D \rvx = \rvf\ \D t + \rmG\ \D \rvw \implies \frac{\D \mSigma_{\text{cov}}}{\D t} = \EE_\rvx[\rvf (\rvx - \vmu)^\trns] + \EE_\rvx[(\rvx - \vmu) \rvf^\trns] + \EE_\rvx[\rmG\rmG^\trns] \\
&\therefore \frac{\D \mSigma_{\GFFDDPM}(t)}{\D t} = \EE_\rvx[-\frac{1}{2}\beta(t)\rvx\rvx^\trns] + \EE_\rvx[\rvx(-\frac{1}{2}\beta(t)\rvx)^\trns] + \EE_\rvx[\sqrt{\beta(t)}\sqrt{\mSigma}\sqrt{\beta(t)}\sqrt{\mSigma}] \\
&\qquad\qquad\qquad = -\beta(t)\mSigma_{\GFFDDPM}(t) + \beta(t)\mSigma = \beta(t)(\mSigma - \mSigma_{\GFFDDPM}(t)) \\
&\implies \frac{\D \mSigma_{\GFFDDPM}(t)}{\mSigma - \mSigma_{\GFFDDPM}(t)} = \beta(t) \D t \implies -\log(\mSigma - \mSigma_{\GFFDDPM}(t)) \vert_{0}^{t} = \int_0^t \beta(s) \D s \\
&\implies -\log(\mSigma - \mSigma_{\GFFDDPM}(t)) + \log(\mSigma - \mSigma_{\rvx}(0)) = \int_0^t \beta(s) \D s \\
&\implies \frac{\mSigma - \mSigma_{\GFFDDPM}(t)}{\mSigma - \mSigma_{\rvx}(0)} = e^{-\int_0^t \beta(s) \D s} \implies \mSigma_{\GFFDDPM}(t) = \mSigma - e^{-\int_0^t \beta(s) \D s} (\mSigma - \mSigma_{\rvx}(0)) \\
&\implies \mSigma_{\GFFDDPM}(t) = \mSigma + e^{-\int_0^t \beta(s) \D s} (\mSigma_{\rvx}(0) - \mSigma)
\end{align*}

For each data point $\rvx(0)$, $\vmu(0) = \rvx(0)$, $\mSigma_{\rvx}(0) = \vzero$:
\begin{align*}
&\implies \vmu_{\GFFDDPM}(t) = \rvx(0)\ e^{-\frac{1}{2}\int_0^t \beta(s) \D s}, \\
&\qquad \quad \mSigma_{\GFFDDPM}(t) = \mSigma + e^{-\int_0^t \beta(s) \D s} (\vzero - \mSigma) = \mSigma(\rmI - \rmI e^{-\int_0^t \beta(s) \D s}) \\
&\therefore p_{0t}(\rvx(t) \mid \rvx(0)) = \gN \left(\rvx(t) \mid \rvx(0)\ e^{-\frac{1}{2}\int_0^t \beta(s) \D s}, \mSigma(\rmI - \rmI e^{-\int_0^t \beta(s) \D s}) \right)
\end{align*}

\newpage
\section{GFF}
\label{GFF}

\begin{figure}[!bth]
\centering
\includegraphics[width=\linewidth]{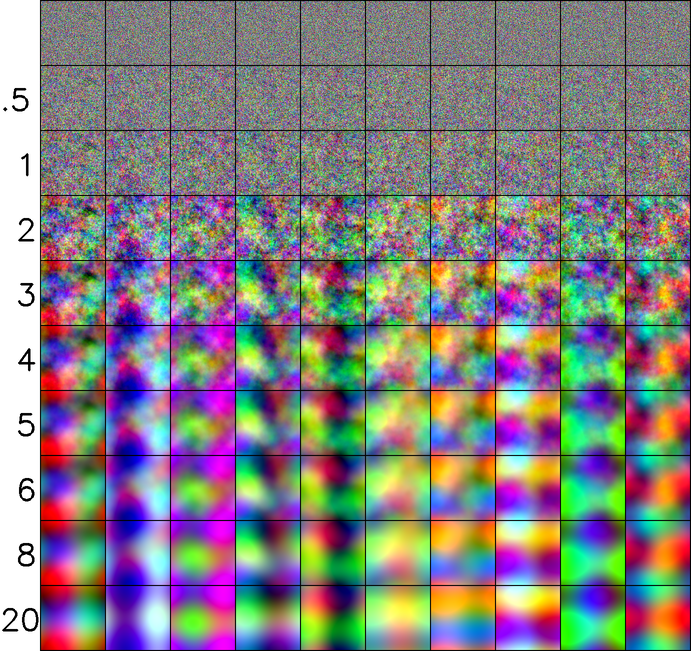}
\caption{(left to right) 10 GFF images (each varying downwards) as a function of the power $\gamma$ of the index (mentioned on the left).}
\label{fig:GFFs}
\end{figure}

A GFF image can be obtained from a noise image in the following way:
\begin{enumerate}
\item
First, sample an $n \x n$ noise image $\rvz$ from the 
standard complex normal distribution with covariance matrix $\Gamma = \rmI_N$ where $N=n^2$ is the total number of pixels, and pseudo-covariance matrix $C = \vzero$: $\rvz \sim \gC \gN(\vzero, \rmI_N, \vzero)$.

The standard \textbf{complex} normal distribution is one where the real part $\rvx$ and imaginary part $\rvy$ are each distributed as the standard normal distribution with variance $\frac{1}{2}\rmI_N$. Let $\mSigma_{\rva \rvb}$ be the covariance matrix between $\rva$ and $\rvb$. We know that $\mSigma_{\rvx \rvx} = \mSigma_{\rvy \rvy} = \frac{1}{2}\rmI_N$, and $\mSigma_{\rvx \rvy} = \mSigma_{\rvy \rvx} = \vzero_N$ Then:
\begin{align}
\Gamma &= \EE_\rvz[\rvz \rvz^\conj] = \mSigma_{\rvx \rvx} + \mSigma_{\rvy \rvy} + i(\mSigma_{\rvy \rvx} - \mSigma_{\rvx \rvy}) = \rmI_N
\label{eq:cov_z}\\
C &= \EE_\rvz[\rvz \rvz^\trns] = \mSigma_{\rvx \rvx} - \mSigma_{\rvy \rvy} + i(\mSigma_{\rvy \rvx} + \mSigma_{\rvx \rvy}) = \vzero_N
\label{eq:psudocov_z}
\end{align}
\item
Apply the Discrete Fourier Transform using the $N \x N$ weights matrix $\rmW_N$: $\rmW_N \rvz$.
\item
Consider a diagonal $N \x N$ matrix of the reciprocal of an index value $k_{ij}$ per pixel $(i,j)$ in Fourier space : $\rmK^{-1} = [1/|k_{ij}|]_{(i,j)}$, and multiply this with the above: $\rmK^{-1} \rmW_N \rvz$.
\item
Take its Inverse Discrete Fourier Transform ($\rmW_N^{-1}$) to make the raw GFF image: $\rmW_N^{-1} \rmK^{-1} \rmW_N \rvz$.
However, this results in a GFF image with a small variance.
\item
Normalize the image with the standard deviation $\sigma_N$ corresponding to its resolution $N$:
\begin{align*}
\rvg &= \frac{1}{\sigma_N}\rmW_N^{-1} \rmK^{-1} \rmW_N \rvz
\numberthis
\label{eq:g}
\quad \iff \quad \rvz = \sigma_N \rmW_N^{-1} \rmK \rmW_N \rvg
\end{align*}
\item
Extract only the real part of $\rvg$, and normalize accordingly (refer \Cref{subsec:GFF_real}):
\begin{align*}
\rvg_{\text{real}} &= \frac{1}{\sqrt{2N}\sigma_N} \Real \left(\rmW_N^{-1} \rmK^{-1} \rmW_N \rvz \right)
\numberthis
\label{eq:g}
\iff \rvz = \sqrt{2N}\sigma_N \Real \left(\rmW_N^{-1} \rmK \rmW_N \rvg_{\text{real}} \right)
\end{align*}
\end{enumerate}

\subsection{Probability distribution of GFF}
\label{subsec:gff_prob}

Let the probability distribution of GFF images be $\mathcal{G}$. This can be seen as a non-isotropic multivariate Gaussian with a non-diagonal covariance matrix $\mSigma$:
\begin{align}
\rvg_\real \sim \gG
= \gN(\vmu, \mSigma)
= \gN(\vzero_N, \mSigma)
\end{align}

We know from the properties of Discrete Fourier Transform (following the normalization convention of the Pytorch / Numpy implementation) that:
\begin{align}
\rmW_N = \rmW_N^\trns; \rmW_N^{-1} = {\rmW_N^{-1}}^\trns; 
\rmW_N^{-1} = \frac{1}{N}\rmW_N^* = \frac{1}{N}\rmW_N^\conj
\label{eq:fourier}
\end{align}

$\vmu$ is given by:
\begin{align*}
\vmu
&= \EE_{\rvg} [ \rvg ]
= \EE_\rvz \left[ \frac{1}{\sigma_N} \rmW_N^{-1} \rmK^{-1} \rmW_N \rvz \right] = \frac{1}{\sigma_N} \rmW_N^{-1} \rmK^{-1} \rmW_N \EE_\rvz \left[ \rvz \right]\\
\implies \vmu &= \vzero_N \qquad [\because \EE_\rvz [\rvz] = \vzero_N]
\numberthis
\label{eq:mu_G}
\end{align*}

$\mSigma$ is given by:
\begin{align*}
\mSigma
&= \EE_{\rvg} [\ \rvg \ \rvg^\trns \ ] \\
&= \EE_\rvz \left[ \frac{1}{\sigma_N}\rmW_N^{-1} \rmK^{-1} \rmW_N \rvz \ \left( \frac{1}{\sigma_N}\rmW_N^{-1} \rmK^{-1} \rmW_N \rvz \right)^\trns \right] \\
&= \EE_\rvz \left[ \frac{1}{\sigma_N^2} \rmW_N^{-1} \rmK^{-1} \rmW_N \rvz \ \rvz^\trns \rmW_N^\trns \rmK^{-1} \rmW_N^{-\trns} \right] \quad \quad [\because \rmK^{-1} \text{ is diagonal }] \\
&= \frac{1}{\sigma_N^2} \rmW_N^{-1} \rmK^{-1} \rmW_N \EE_\rvz [\rvz \rvz^\trns] \rmW_N \rmK^{-1} \rmW_N^{-1}\\
&= \frac{1}{\sigma_N^2} \rmW_N^{-1} \rmK^{-1} \rmW_N \rmW_N \rmK^{-1} \rmW_N^{-1} \qquad \qquad \qquad [\because \EE_\rvz [\rvz \rvz^\trns] = \rmI_N] \\
&= \left( \frac{1}{\sigma_N}\rmW_N^{-1} \rmK^{-1} \rmW_N \right) \left( \frac{1}{\sigma_N}\rmW_N^{-1} \rmK^{-1} \rmW_N \right)^\trns \quad [\because (\rmW_N^{-1} \rmK^{-1} \rmW_N)^\trns = \rmW_N \rmK^{-1} \rmW_N^{-1}]
% \\
% &= \left( \frac{1}{\sigma_N}\rmW_N^{-1} \rmK^{-1} \rmW_N \right) \left( \frac{1}{\sigma_N}\rmW_N^{-1} \rmK^{-1} \rmW_N \right) \quad [\text{for the real component}] \\
% &= \frac{1}{\sigma_N^2}\rmW_N^{-1} \rmK^{-1} \rmK^{-1} \rmW_N
\end{align*}

% Hence, $\rvg_\real \sim \gN(\vzero, \mSigma)$, where $\mSigma$ is given in \cref{eq:covG}.

$\sigma_N^2$ is such that the variance of each variable is 1, i.e. each diagonal element of $\mSigma$ is 1. Thus,
\begin{align*}
\sigma_N^2 = \Var[\mSigma_N] \quad \text{where } \mSigma_N = \rmW_N^{-1} \rmK^{-1} \rmW_N \rmW_N \rmK^{-1} \rmW_N^{-1}
\numberthis
\end{align*}
Hence,
\begin{align*}
\sqrt{\mSigma} &= \frac{1}{\sigma_N}\rmW_N^{-1} \rmK^{-1} \rmW_N,
\numberthis
\label{eq:covG_sqrt2}
\\
\mSigma &= \sqrt{\mSigma} \sqrt{\mSigma}^\trns ,
\numberthis
\label{eq:covG2}
\\
\sqrt{\mSigma^{-1}} &= \sigma_N \rmW_N^{-1} \rmK \rmW_N
\numberthis
\label{eq:covGinv_sqrt2}
\\
\mSigma^{-1} &= \sqrt{\mSigma^{-1}}^\trns \sqrt{\mSigma^{-1}},
\numberthis
\label{eq:covGinv2}
\end{align*}

\subsection{Derivation of GFF \texorpdfstring{$\mSigma$}{Covariance} for real \texorpdfstring{$\rvg$}{g} and complex \texorpdfstring{$\rvz$}{z}}
\label{subsec:GFF_real}

\textbf{z is complex, g is real} :
$\vmu$ is given by:
\begin{align*}
\vmu
&= \EE_{\rvg_\real} [ \rvg_\real ]
= \EE_\rvg [\ \frac{1}{2\sigma_N}(\rvg + \rvg^*) \ ]
= \frac{1}{2\sigma_N} \left( \EE_\rvg [\rvg] + \EE_\rvg [\rvg^*] \right)\\
&= \frac{1}{2\sigma_N} \left( \EE_\rvz [\rmW_N^{-1} \rmK^{-1} \rvz] + \EE_\rvz [(\rmW_N^{-1} \rmK^{-1} \rvz)^*] \right)\\
&= \frac{1}{2\sigma_N} \left( \rmW_N^{-1} \rmK^{-1} \EE_\rvz [\rvz] + \rmW_N^{-1 *} \rmK^{-1} \EE_\rvz [\rvz^*] \right)\\
\implies \vmu &= \vzero_N \qquad \qquad [\because \EE_\rvz [\rvz] = \EE_\rvz [\rvz^*] = \vzero_N]
\numberthis
\label{eq:mu_G}
\end{align*}

\textbf{z is complex, g is real} :
$\mSigma$ is given by:
\begin{align*}
\mSigma
&= \EE_{\rvg_\real} [\ \rvg_\real \ \rvg_\real^\trns \ ] \\
&= \EE_\rvg [\ \frac{1}{2\sigma_N}(\rvg + \rvg^*) \ \frac{1}{2\sigma_N}(\rvg + \rvg^*)^\trns \ ] \\
&= \frac{1}{4\sigma_N^2} \EE_\rvg [\ (\rvg + \rvg^*) \ (\rvg^\trns + \rvg^\conj) \ ] \\
&= \frac{1}{4\sigma_N^2} \EE_\rvg [\ \rvg \rvg^\trns + \rvg \rvg^\conj + \rvg^* \rvg^\trns + \rvg^* \rvg^\conj \ ] \\
&= \frac{1}{4\sigma_N^2}\left( \EE_\rvg [\rvg \rvg^\trns] + \EE_\rvg [\rvg \rvg^\conj] + \EE_\rvg [\rvg^* \rvg^\trns] + \EE_\rvg [\rvg^* \rvg^\conj] \right)
\\
\EE_\rvg [\rvg \rvg^\trns]
&= \EE_\rvz [\rmW_N^{-1} \rmK^{-1} \rvz \ (\rmW_N^{-1} \rmK^{-1} \rvz)^\trns] \\
&= \EE_\rvz [\rmW_N^{-1} \rmK^{-1} \rvz \ \rvz^\trns \rmK^{-1} \rmW_N^{-\trns} ] \quad [\because \rmK^{-1} \text{ is diagonal }] \\
&= \rmW_N^{-1} \rmK^{-1} \EE_\rvz [\rvz \rvz^\trns] \rmK^{-1} \rmW_N^{-T}\\
&= \vzero_N \qquad \qquad \qquad \qquad \qquad \qquad \quad [\because \EE_\rvz [\rvz \rvz^\trns] = \vzero_N \text{ (\cref{eq:psudocov_z})}]
\\
\EE_\rvg [\rvg \rvg^\conj]
&= \EE_\rvz [\rmW_N^{-1} \rmK^{-1} \rvz \ (\rmW_N^{-1} \rmK^{-1} \rvz)^\conj] \\
&= \EE_\rvz [\rmW_N^{-1} \rmK^{-1} \rvz \ \rvz^\conj \rmK^{-1} \rmW_N^{-\conj} ] \quad [\because \mK^{-1} \text{ is real diagonal }] \\
&= \rmW_N^{-1} \rmK^{-1} \EE_\rvz [\rvz \rvz^\conj] \rmK^{-1} \frac{1}{N}\rmW_N  \quad [\because \rmW_N^{-1} = \frac{1}{N}\rmW_N^\conj \text{ (\cref{eq:fourier})} ] \\
&= \frac{1}{N} \rmW_N^{-1} \rmK^{-1} \rmK^{-1} \rmW_N \qquad \qquad [\because \EE_\rvz [\rvz \rvz^\conj] = \rmI_N \text{ (\cref{eq:cov_z})}]
\\
\EE_\rvg [\rvg^* \rvg^\trns]
&= \EE_\rvz [(\rmW_N^{-1} \rmK^{-1} \rvz)^* \ (\rmW_N^{-1} \rmK^{-1} \rvz)^\trns] \\
&= \EE_\rvz [\rmW_N^{-1 *} \rmK^{-1} \rvz^* \ \rvz^\trns \rmK^{-1} \rmW_N^{-1} ] \\
&= \frac{1}{N}\rmW_N \rmK^{-1} \EE_\rvz [\rvz^* \rvz^\trns] \rmK^{-1} \rmW_N^{-1} \quad [\because \rmW_N^{-1} = \frac{1}{N}\rmW_N^* \text{ (\cref{eq:fourier})} ] \\
&= \frac{1}{N} \rmW_N \rmK^{-1} \rmK^{-1} \rmW_N^{-1} \quad [\because \EE_\rvz [\rvz^* \rvz^\trns] = \EE_\rvz [\rvz \rvz^\conj]^* = \rmI_N \text{ (\cref{eq:cov_z})}]
\\
\EE_\rvg [\rvg^* \rvg^\conj]
&= \EE_\rvz [(\rmW_N^{-1} \rmK^{-1} \rvz)^* \ (\rmW_N^{-1} \rmK^{-1} \rvz)^\conj] \\
&= \EE_\rvz [\rmW_N^{-1 *} \rmK^{-1} \rvz^* \ \rvz^\conj \rmK^{-1} \rmW_N^{-\conj} ] \\
&= \rmW_N^{-1 *} \rmK^{-1} \EE_\rvz [\rvz^* \rvz^\conj] \rmK^{-1} \rmW_N^{-\conj} \\
&= \vzero_N \qquad \qquad \qquad \qquad [\because \EE_\rvz [\rvz^* \rvz^\conj] = \EE_\rvz [\rvz \rvz^\trns]^* = \vzero_N \text{ (\cref{eq:cov_z})}]
\\
\implies \mSigma &= \frac{1}{4\sigma_N^2}\left( \vzero + \frac{1}{N} \rmW_N^{-1} \rmK^{-1} \rmK^{-1} \rmW_N + \frac{1}{N} \rmW_N \rmK^{-1} \rmK^{-1} \rmW_N^{-1} + \vzero \right)\\
&= \frac{1}{4N\sigma_N^2}\left(\rmW_N^{-1} \rmK^{-1} \rmK^{-1} \rmW_N + \rmW_N \rmK^{-1} \rmK^{-1} \rmW_N^{-1} \right)\\
&= \frac{1}{4N\sigma_N^2}\left(\rmW_N^{-1} \rmK^{-1} \rmK^{-1} \rmW_N + (N \rmW_N^{-1 *}) \rmK^{-1} \rmK^{-1} (\frac{1}{N}\rmW_N^*) \right)\\
&= \frac{1}{2N\sigma_N^2}\left(\frac{1}{2}\left(\rmW_N^{-1} \rmK^{-1} \rmK^{-1} \rmW_N + (\rmW_N^{-1} \rmK^{-1} \rmK^{-1} \rmW_N)^*\right) \right)\\
\implies \mSigma &= \frac{1}{2N\sigma_N^2} \Real \left( \rmW_N^{-1} \rmK^{-1} \rmK^{-1} \rmW_N \right),
\numberthis
\label{eq:covG}
\\
\sqrt{\mSigma} &= \frac{1}{\sqrt{2N}\sigma_N} \Real \left( \rmW_N^{-1} \rmK^{-1} \rmW_N \right),
\numberthis
\label{eq:covG_sqrt}
\\
\mSigma^{-1} &= 2N\sigma_N^2 \Real \left( \rmW_N^{-1} \rmK \rmK \rmW_N \right),
\numberthis
\label{eq:covGinv}
\\
\sqrt{\mSigma^{-1}} &= \sqrt{2N}\sigma_N \Real \left( \rmW_N^{-1} \rmK \rmW_N \right)
\numberthis
\label{eq:covGinv_sqrt}
\end{align*}

\subsection{Log probability of transformation}
\label{subsec:gff_logprob}
\begin{align*}
\rvg &= \sqrt{\mSigma}\rvz \\
\implies \log p(\rvg) &= \log p(\rvz) - \log \Abs{\det \frac{\D \rvg}{\D \rvz}} \\
&= \log p(\rvz) - \log \Abs{\det \sqrt{\mSigma}} \\
&= \log p(\rvz) - \log \Abs{\det \frac{1}{\sqrt{2N}\sigma_N}\Real\left(\rmW_N^{-1} \rmK^{-1} \rmW_N\right) } \\
&= \log p(\rvz) - \frac{1}{\sqrt{2N}\sigma_N} \log \Abs{\det\rmK^{-1} }
\end{align*}

This is useful for building (normalizing) flows using non-isotropic Gaussian noise.

\subsection{Varying $\rmK$}

% \begin{figure}[!tb]
% \centering
% \includegraphics[width=\linewidth]{figs/GFFs.png}
% \caption{10 GFF images (left to right) as a function of the power $\gamma$ of the index (specified on the left, varying downwards). The top row is $\gamma=0$ i.e. isotropic Gaussian. It can be seen that correlation between neighbouring pixel values increases with increase in $\gamma$.}
% \label{fig:GFFs}
% \end{figure}

The index matrix $\rmK$ involves computation of an index value $k_{ij}$ per pixel $(i,j)$. However, this index value could be raised to any power $\gamma$ i.e. $|k_{ij}|^\gamma$. The effect of varying $\gamma$ can be seen in \Cref{fig:GFFs} : greater the $\gamma$, the more correlated are neighbouring pixels.

\newpage
\section{Score Matching Langevin Dynamics (SMLD) \cite{song2019generative, song2020improved}}
\label{sec:smld}

\subsection{Score for SMLD}

For isotropic Gaussian noise as in SMLD,
\begin{align}
&q_{\sigma_t}^\SMLD (\rvx_i \mid \rvx) = \gN(\rvx_i \mid \rvx, \sigma_i^2 \rmI) \implies \rvx_i = \rvx + \sigma_i \rvepsilon \\
&\implies \nabla_{\rvx_i} \log q_{\sigma_i}^\SMLD (\rvx_i \mid \rvx) = -\frac{1}{\sigma_i^2}(\rvx_i - \rvx) = -\frac{1}{\sigma_i}\rvepsilon
% \\
% &q_{\sigma_i}^\SMLD (\rvx_{i+1} \mid \rvx_i) = \gN(\rvx_{i+1} \mid \rvx_i, (\sigma_{i+1}^2 - \sigma_i^2)\rmI) \\
% &\implies \rvx_i = \rvx_{i-1} + \sqrt{\sigma_i^2 - \sigma_{i-1}^2}\rvepsilon_{i-1}
 \end{align}

\subsection{Objective function for SMLD}

The objective function for SMLD at noise level $\sigma$ is:
\begin{align}
&\loss^\SMLD(\rvtheta; \sigma_i) \triangleq\ \frac{1}{2} \EE_{q_{\sigma_i}(\rvx_i \mid \rvx) p(\rvx)} \bigg[ \Norm{\rvs_\rvtheta (\rvx_i, \sigma_i) + \frac{1}{\sigma_i^2} (\rvx_i - \rvx) }_2^2 \bigg]
\end{align}

\subsection{Variance of actual score for SMLD}
\begin{align}
\EE\left[ \Norm{ \nabla_{\rvx_i} \log q_{\sigma_i}^\SMLD (\rvx_i \mid \rvx) }_2^2 \right] &= \EE\left[ \Norm{ -\frac{(\rvx_i - \rvx)}{\sigma_i^2} }_2^2 \right] = \EE\left[ \Norm{ \frac{\sigma_i\rvepsilon}{\sigma_i^2} }_2^2 \right] = \frac{1}{\sigma_i^2}\EE\left[ \Norm{\rvepsilon}_2^2 \right] = \frac{1}{\sigma_i^2}
\end{align}

\subsection{Overall objective function for SMLD}

\cite{song2019generative, song2020improved} chose a geometric series of $\sigma_i$'s, i.e. $\sigma_{i-1}/\sigma_i = \gamma$. The overall objective function was a weighted combination of the objectives at different noise levels, the weight being $\lambda(\sigma_i) = \sigma_i^2$:
\begin{align*}
\gL^\SMLD(\rvtheta; \{\sigma_i\}_{i=1}^{L}) &\triangleq \frac{1}{2L} \sum_{i=1}^{L} \EE_{q_{\sigma_i}(\rvx_i \mid \rvx) p(\rvx)} \bigg[ \Norm{\sigma_i \rvs_\rvtheta (\rvx_i, \sigma_i) + \frac{(\tilde \rvx - \rvx)}{\sigma_i} }_2^2 \bigg] \\
&= \frac{1}{2L} \sum_{i=1}^{L} \EE_{q_{\sigma_i}(\rvx_i \mid \rvx) p(\rvx)} \bigg[ \Norm{\sigma_i \rvs_\rvtheta (\rvx_i, \sigma_i) + \rvepsilon }_2^2 \bigg]
\numberthis
\end{align*}

\subsection{Unconditional SMLD score estimation}

Song et. al. discovered that empirically the estimated score was proportional to $\frac{1}{\sigma}$. So an unconditional score model is:
\begin{align}
\rvs_\rvtheta(\rvx_i, \sigma_i) = -\frac{1}{\sigma_i} \rvepsilon_\rvtheta(\rvx_i)
\end{align}

In this case, the overall objective function changes to:
\begin{align*}
\gL^\SMLD(\rvtheta; \{\sigma_i\}_{i=1}^{L}) &\triangleq \frac{1}{2L} \sum_{i=1}^{L} \EE_{q_{\sigma_i}(\rvx_i \mid \rvx) p(\rvx)} \bigg[ \Norm{ \rvepsilon - \rvepsilon_\rvtheta (\rvx_i) }_2^2 \bigg]
\numberthis
\\
&= \frac{1}{2L} \sum_{i=1}^{L} \EE_{q_{\sigma_i}(\rvx_i \mid \rvx) p(\rvx)} \bigg[ \Norm{ \rvepsilon - \rvepsilon_\rvtheta (\rvx + \sigma_i \rvepsilon) }_2^2 \bigg]
\end{align*}

\subsection{Sampling in SMLD}

$i=0$ corresponds to data, and $i=L$ corresponds to noise. Hence, $i = L, \cdots, 0$ is the time order for sampling.

% Forward : $\rvx_i = \rvx_{i-1} + \sqrt{\sigma_i^2 - \sigma_{i-1}^2}\rvepsilon_{i-1}$

% \underline{Reverse}:

Using ALS from ~\cite{song2019generative, song2020improved}:
\begin{align*}
&\rvx_L^M \sim \gN(\rvx \mid \vzero, \sigma_{\max}\rmI)\\
&\begin{rcases}
&\rvx_i^M = \rvx_{i+1}^0 \\
&\alpha_i = \eps \sigma_i^2/\sigma_{\min}^2 \\
&\rvx_i^{m-1} \leftarrow \rvx_i^m + \alpha_i \rvs_{\rvtheta^*}(\rvx_i^m, \sigma_i) + \sqrt{2\alpha_i} \rvepsilon_i^{m-1} , m=M,\cdots,0\\
\implies &\rvx_i^{m-1} \leftarrow \rvx_i^m - \frac{\alpha_i}{\sigma_i} \rvepsilon_{\rvtheta^*}(\rvx_i^m) + \sqrt{2\alpha_i} \rvepsilon_i^{m-1} , m=M,\cdots,0
\end{rcases}
i = L, \cdots, 1
\numberthis
\end{align*}

Using Consistent Annealed Sampling~\cite{jolicoeur2020adversarial}:
\begin{align*}
&\alpha_i = \eps \sigma_i^2/\sigma_{\min}^2 = \eta \sigma_i^2;\ \beta = \sqrt{1 - \gamma^2(1 - \eps/\sigma_{\min}^2)^2};\ \gamma = \sigma_i/\sigma_{i-1} ; \sigma_i > \sigma_{i-1} \\
&\rvx_{i-1} \leftarrow \rvx_i + \alpha_i \rvs_{\rvtheta^*}(\rvx_i, \sigma_i) + \beta \sigma_{i-1} \rvepsilon_{i-1}, i = L, \cdots, 1 \\
\implies &\rvx_{i-1} \leftarrow \rvx_i - \eta \sigma_i \rvepsilon_{\rvtheta^*}(\rvx_i) + \beta \sigma_{i-1} \rvepsilon_{i-1}, i = L, \cdots, 1
\numberthis
\end{align*}

\subsection{Expected Denoised Sample}

From \cite{saremi2019neb}, assuming isotropic Gaussian noise, we know that the expected denoised sample $\rvx^*(\rvx_i, \sigma_i) \triangleq \EE_{\rvx \sim q_{\sigma_i}(\rvx \mid \rvx_i)}[\rvx]$ and the optimal score $\rvs_{\rvtheta^*}(\rvx_i, \sigma_i)$ are related as:
\begin{align*}
&\rvs_{\rvtheta^*}(\rvx_i, \sigma_i) = \frac{1}{\sigma_i^2}(\rvx^*(\rvx_i, \sigma_i) - \rvx_i) \\
\implies &\rvx^*(\rvx_i, \sigma_i) = \rvx_i + \sigma_i^2 \rvs_{\rvtheta^*}(\rvx_i, \sigma_i) = \rvx_i - \sigma_i \rvepsilon_{\rvtheta^*}(\rvx_i)
\numberthis
\end{align*}

\subsection{SDE formulation : Variance Exploding (VE) SDE}

Forward process:
\begin{align*}
\rvx_i &= \rvx_{i-1} + \sqrt{\sigma_i^2 - \sigma_{i-1}^2}\rvepsilon_{i-1} \\
\implies \rvx(t + \Delta t) &= \rvx(t) + \sqrt{(\sigma^2(t + \Delta t) - \sigma^2(t)) \Delta t}\ \rvepsilon(t) \\
&\approx \rvx(t) + \sqrt{\frac{\D [\sigma^2(t)]}{\D t} \Delta t}\ \rvw(t) \\
\implies \D \rvx &= \sqrt{\frac{\D [\sigma^2(t)]}{\D t}}\ \D \rvw
\numberthis
% \\
% \sigma_{i}/\sigma_{i+1} &= \gamma \implies \sigma(i) = \sigma_L \gamma^{L-i} \implies \sigma(t) \approx \sigma_L \gamma^{L(1-t)} \\ \implies \frac{\D \sigma^2(t)}{\D t} &= \frac{\D \sigma_L^2 \gamma^{2(L-t)}}{\D t} = \sigma_L^2 (2(L-t)\gamma^{2(L-t) - 1})
\end{align*}

\begin{align*}
&\D \rvx = \rvf(\rvx, t) \D t + \rmL(\rvx, t) \D \rvw
\implies \frac{\D \vmu}{\D t} = \EE_\rvx[\rvf(\rvx, t)], \\
&\frac{\D \textbf{Cov}[\rvx]}{\D t} = \EE_\rvx[\rvf(\rvx, t) (\rvx - \vmu)^\trns] + \EE_\rvx[(\rvx - \vmu) \rvf(\rvx, t)^\trns] + \EE_\rvx[\rmL(\rvx, t) \rmQ \rmL^\trns(\rvx, t)]
\end{align*}
where $\rvw$ is Brownian motion, $\rmQ$ is the PSD of $\rvw$. For GFF noise, $\rmQ = \mSigma$.

Mean and Covariance:
\begin{align*}
&\frac{\D \vmu_{\SMLD}(t)}{\D t} = \vzero \implies \vmu_{\SMLD}(t) = \vmu(0) \\
&\frac{\D \mSigma_{\SMLD}(t)}{\D t} = \EE_\rvx\left[\sqrt{\frac{\D [\sigma^2(t)]}{\D t}} \sqrt{\frac{\D [\sigma^2(t)]}{\D t}}\right] = \frac{\D [\sigma^2(t)]}{\D t}
\end{align*}

\newpage
\section{Non-isotropic SMLD (NI-SMLD)}
\label{sec:nismld}

\subsection{Score for NI-SMLD}

\begin{align}
% &q_{\sigma_i}^{\NI-\SMLD} (\rvx_{i+1} \mid \rvx_i) = \gN(\rvx_{i+1} \mid \rvx_i, (\sigma_{i+1}^2 - \sigma_i^2)\mSigma) \\
% &\implies \rvx_i = \rvx_{i-1} + \sqrt{\sigma_i^2 - \sigma_{i-1}^2}\sqrt{\mSigma}\rvepsilon_{i-1} \\
&q_{\sigma_i}^\SMLD (\rvx_i \mid \rvx) = \gN(\rvx_i \mid \rvx, \sigma_i^2 \mSigma) \implies \rvx_i = \rvx + \sigma_i \sqrt{\mSigma} \rvepsilon \implies \rvepsilon = \sqrt{\mSigma^{-1}}\frac{\rvx_i - \rvx}{\sigma_i} \\
&\implies \nabla_{\rvx_i} \log q_{\sigma_i}^\SMLD (\rvx_i \mid \rvx) = -\mSigma^{-1}\frac{\rvx_i - \rvx}{\sigma_i^2} = -\sqrt{\mSigma^{-1}}\frac{\rvepsilon}{\sigma_i}
\end{align}

\subsection{Objective function for NI-SMLD}

The objective function for SMLD at noise level $\sigma$ is:
\begin{align}
\loss^{\NI-\SMLD}(\rvtheta; \sigma_i) &\triangleq\ \frac{1}{2} \EE_{q_{\sigma_i}(\rvx_i \mid \rvx) p(\rvx)} \bigg[ \Norm{\rvs_\rvtheta (\rvx_i, \sigma_i) + \mSigma^{-1}\frac{\rvx_i - \rvx}{\sigma_i^2} }_2^2 \bigg] \\
&=\ \frac{1}{2} \EE_{q_{\sigma_i}(\rvx_i \mid \rvx) p(\rvx)} \bigg[ \Norm{\rvs_\rvtheta (\rvx_i, \sigma_i) + \frac{1}{\sigma_i}\sqrt{\mSigma^{-1}}\rvepsilon }_2^2 \bigg]
\end{align}

\subsection{Expected value of score for NI-SMLD}
\begin{align*}
&\EE\left[ \Norm{ \nabla_{\rvx_i} \log q_{\sigma_i}^{\NI-\SMLD} (\rvx_i \mid \rvx) }_2^2 \right] = \EE\left[ \Norm{ -\mSigma^{-1}\frac{\rvx_i - \rvx}{\sigma_i^2} }_2^2 \right] \\
&= \EE\left[ \Norm{ \mSigma^{-1} \frac{\sigma_i\sqrt{\mSigma}\rvepsilon}{\sigma_i^2} }_2^2 \right] = \frac{1}{\sigma_i^2}\mSigma^{-1}\EE\left[ \Norm{\rvepsilon}_2^2 \right] = \frac{1}{\sigma_i^2} \mSigma^{-1} \dim(\rvepsilon)
\numberthis
\end{align*}

\subsection{Overall objective function for NI-SMLD}

$\implies \lambda(\sigma_i) = \sigma_i^2\mSigma$
\begin{align*}
\gL^{\NI-\SMLD}(\rvtheta; \{\sigma_i\}_{i=1}^{L}) 
&\triangleq \frac{1}{2L} \sum_{i=1}^{L} \EE_{q_{\sigma_i}(\rvx_i \mid \rvx) p(\rvx)} \bigg[ \Norm{\sigma_i \sqrt{\mSigma} \rvs_\rvtheta (\rvx_i, \sigma_i) + \sqrt{\mSigma^{-1}} \frac{(\tilde \rvx - \rvx)}{\sigma_i} }_2^2 \bigg] \\
&= \frac{1}{2L} \sum_{i=1}^{L} \EE_{q_{\sigma_i}(\rvx_i \mid \rvx) p(\rvx)} \bigg[ \Norm{\sigma_i \sqrt{\mSigma} \rvs_\rvtheta (\rvx_i, \sigma_i) + \rvepsilon }_2^2 \bigg]
\numberthis
\end{align*}

\subsection{Unconditional NI-SMLD score estimation}

An unconditional score model is:
\begin{align}
\rvs_\rvtheta(\rvx_i, \sigma_i) = -\sqrt{\mSigma^{-1}} \frac{1}{\sigma_i} \rvepsilon_\rvtheta(\rvx_i)
\end{align}

In this case, the overall objective function changes to:
\begin{align*}
\gL^{\GFF-\SMLD}(\rvtheta; \{\sigma_i\}_{i=1}^{L}) 
&\triangleq \frac{1}{2L} \sum_{i=1}^{L} \EE_{q_{\sigma_i}(\rvx_i \mid \rvx) p(\rvx)} \bigg[ \Norm{ \rvepsilon - \rvepsilon_\rvtheta (\rvx_i) }_2^2 \bigg] \\
&= \frac{1}{2L} \sum_{i=1}^{L} \EE_{q_{\sigma_i}(\rvx_i \mid \rvx) p(\rvx)} \bigg[ \Norm{ \rvepsilon - \rvepsilon_\rvtheta (\rvx + \sigma_i\sqrt{\mSigma}\rvepsilon) }_2^2 \bigg]
\numberthis
\end{align*}

\subsection{Sampling in NI-SMLD}

$i=0$ corresponds to data, and $i=L$ corresponds to noise. Hence, $i = L, \cdots, 0$ is the time order for sampling.

Forward : $\rvx_i = \rvx_{i-1} + \sqrt{\sigma_i^2 - \sigma_{i-1}^2}\sqrt{\mSigma}\rvepsilon_{i-1}$

Reverse:

From Song et. al., ALS:
\begin{align*}
&\rvx_L^0 \sim \gN(\rvx \mid \vzero, \sigma_{\max} \sqrt{\mSigma})\\
&\begin{rcases}
&\rvx_i^0 = \rvx_{i+1}^M
\\
&\rvx_i^{m+1} \leftarrow \rvx_i^m + \alpha_i \rvs_{\rvtheta^*}(\rvx_i^m, \sigma_i) + \sqrt{2\alpha_i} \sqrt{\mSigma} \rvepsilon_i^{m+1} , m=1,\cdots,M
\end{rcases}
i = L, \cdots, 1
\numberthis
\\
&\alpha_i = \eps \sigma_i^2/\sigma_L^2
\end{align*}

From Alexia et. al., CAS:
\begin{align*}
&\rvx_{i-1} \leftarrow \rvx_i + \alpha_i \rvs_{\rvtheta^*}(\rvx_i, \sigma_i) + \beta \sigma_{i-1} \sqrt{\mSigma} \rvepsilon_{i-1}, i = L, \cdots, 1
\numberthis
\\
&\alpha_i = \eps \sigma_t^2/\sigma_{\min}^2;\ \beta = \sqrt{1 - \gamma^2(1 - \eps/\sigma_{\min}^2)^2};\ \gamma = \sigma_t/\sigma_{t-1} ; \sigma_t > \sigma_{t-1}
\end{align*}

\subsection{Expected Denoised Sample}

From \cite{saremi2019neb}, assuming isotropic Gaussian noise of covariance $\sigma^2\mSigma$, we know that the expected denoised sample $\rvx^*(\tilde\rvx, \sigma) \triangleq \EE_{\rvx \sim q_\sigma(\rvx \mid \tilde\rvx)}[\rvx]$ and the optimal score $\rvs_{\rvtheta^*}(\tilde\rvx, \sigma)$ are related as:
\begin{align*}
&\rvs_{\rvtheta^*}(\tilde\rvx, \sigma) = \frac{1}{\sigma^2}\mSigma^{-1}(\rvx^*(\tilde\rvx, \sigma) - \tilde\rvx) \\
\implies &\rvx^*(\tilde\rvx, \sigma) = \tilde\rvx + \sigma^2 \mSigma\ \rvs_{\rvtheta^*}(\tilde\rvx, \sigma) = \tilde\rvx - \sigma \sqrt{\mSigma}\ \rvepsilon_{\rvtheta^*}(\tilde\rvx)
\numberthis
\end{align*}

\subsection{Initial noise scale for NI-SMLD}

Let $\hat{p}_{\sigma_{1}}(\rvx) \triangleq \frac{1}{N} \sum_{i=1}^N p^{(i)}(\rvx)$, where $p^{(i)}(\rvx) \triangleq \gN(\rvx \mid \rvx^{(i)}, \sigma_1^2 \mSigma)$. With $r^{(i)}(\rvx) \triangleq \frac{p^{(i)}(\rvx)}{\sum_{k=1}^N  p^{(k)}(\rvx) }$, the score function is $\nabla_\rvx \log \hat{p}_{\sigma_{1}}(\rvx) = \sum_{i=1}^N r^{(i)}(\rvx) \nabla_\rvx \log p^{(i)}(\rvx)$.

We know that:
\begin{align*}
\gN(\rvx \mid \rvx^{(i)}, \sigma_1^2 \mSigma) = \frac{1}{(2\pi)^{D/2} \sigma_1^D \abs{\mSigma}^{1/2}} \exp \left( -\frac{1}{2\sigma_1^2} (\rvx - \rvx^{(i)})^\trns \mSigma^{-1} (\rvx - \rvx^{(i)}) \right)
\end{align*}

\begin{align*}
    &\EE_{p^{(i)}(\rvx)}[r^{(j)}(\rvx)] = \int \frac{p^{(i)}(\rvx)p^{(j)}(\rvx)}{\sum_{k=1}^N p^{(k)}(\rvx)} \ud \rvx \leq \int \frac{p^{(i)}(\rvx)p^{(j)}(\rvx)}{p^{(i)}(\rvx) + p^{(j)}(\rvx)} \ud \rvx \\
    &= \frac{1}{2} \int \frac{2}{\frac{1}{p^{(i)}(\rvx)} + \frac{1}{p^{(j)}(\rvx)}} \ud \rvx \leq \frac{1}{2} \int \sqrt{p^{(i)}(\rvx) p^{(j)}(\rvx)} \ud \rvx\\
    &= \frac{1}{2} \frac{1}{(2\pi)^{D/2}\sigma_1^D \abs{\mSigma}^{1/2}} \int \exp \bigg(-\frac{1}{4\sigma_1^2}\bigg( (\rvx - \rvx^{(i)})^\trns \mSigma^{-1} (\rvx - \rvx^{(i)}) + (\rvx - \rvx^{(j)})^\trns \mSigma^{-1} (\rvx - \rvx^{(j)}) \bigg) \ud \rvx\\
    &= \frac{1}{2} \frac{1}{(2\pi)^{D/2}\sigma_1^D \abs{\mSigma}^{1/2}} \int \exp \bigg(-\frac{1}{4\sigma_1^2}\bigg( (\rvx - \rvx^{(i)})^\trns \mSigma^{-1} (\rvx - \rvx^{(i)}) + (\rvx - \rvx^{(j)})^\trns \mSigma^{-1} (\rvx - \rvx^{(i)} + \rvx^{(i)} - \rvx^{(j)}) \bigg) \ud \rvx\\
    &= \frac{1}{2} \frac{1}{(2\pi)^{D/2}\sigma_1^D \abs{\mSigma}^{1/2}} \int \exp \bigg(-\frac{1}{4\sigma_1^2}\bigg( (\rvx - \rvx^{(i)})^\trns \mSigma^{-1} (\rvx - \rvx^{(i)}) + (\rvx - \rvx^{(j)})^\trns \mSigma^{-1} (\rvx - \rvx^{(i)})\\
    &\qquad \qquad \qquad \qquad \qquad \qquad \qquad + (\rvx - \rvx^{(j)})^\trns \mSigma^{-1} (\rvx^{(i)} - \rvx^{(j)}) \bigg) \ud \rvx\\
    &= \frac{1}{2} \frac{1}{(2\pi)^{D/2}\sigma_1^D \abs{\mSigma}^{1/2}} \int \exp \bigg(-\frac{1}{4\sigma_1^2}\bigg( (\rvx - \rvx^{(i)})^\trns \mSigma^{-1} (\rvx - \rvx^{(i)}) + (\rvx - \rvx^{(i)} + \rvx^{(i)} - \rvx^{(j)})^\trns \mSigma^{-1} (\rvx - \rvx^{(i)})\\
    &\qquad \qquad \qquad \qquad \qquad \qquad \qquad + (\rvx - \rvx^{(i)} + \rvx^{(i)} - \rvx^{(j)})^\trns \mSigma^{-1} (\rvx^{(i)} - \rvx^{(j)}) \bigg) \ud \rvx\\
    &= \frac{1}{2} \frac{1}{(2\pi)^{D/2}\sigma_1^D \abs{\mSigma}^{1/2}} \int \exp \bigg(-\frac{1}{4\sigma_1^2}\bigg( (\rvx - \rvx^{(i)})^\trns \mSigma^{-1} (\rvx - \rvx^{(i)}) + (\rvx - \rvx^{(i)})^\trns \mSigma^{-1} (\rvx - \rvx^{(i)}) \\
    &\qquad \qquad \qquad \qquad \qquad \qquad \qquad + (\rvx^{(i)} - \rvx^{(j)})^\trns \mSigma^{-1} (\rvx - \rvx^{(i)}) + (\rvx - \rvx^{(i)})^\trns \mSigma^{-1} (\rvx^{(i)} - \rvx^{(j)}) \\
    &\qquad \qquad \qquad \qquad \qquad \qquad \qquad + (\rvx^{(i)} - \rvx^{(j)})^\trns \mSigma^{-1} (\rvx^{(i)} - \rvx^{(j)}) \bigg) \ud \rvx\\
    &= \frac{1}{2} \frac{1}{(2\pi)^{D/2}\sigma_1^D \abs{\mSigma}^{1/2}} \int \exp \bigg(-\frac{1}{2\sigma_1^2}\bigg( (\rvx - \rvx^{(i)})^\trns \mSigma^{-1} (\rvx - \rvx^{(i)}) \\
    &\qquad \qquad \qquad \qquad \qquad \qquad \qquad + (\rvx - \rvx^{(i)})^\trns \mSigma^{-1} (\rvx^{(i)} - \rvx^{(j)}) \\
    &\qquad \qquad \qquad \qquad \qquad \qquad \qquad + \frac{1}{2}(\rvx^{(i)} - \rvx^{(j)})^\trns \mSigma^{-1} (\rvx^{(i)} - \rvx^{(j)}) \bigg) \ud \rvx\\
    &= \frac{1}{2} \frac{1}{(2\pi)^{D/2}\sigma_1^D \abs{\mSigma}^{1/2}} \int \exp \bigg(-\frac{1}{2\sigma_1^2}\bigg( (\rvx - \rvx^{(i)})^\trns \mSigma^{-1} (\rvx - \rvx^{(i)}) \\
    &\qquad \qquad \qquad \qquad \qquad \qquad \qquad + 2(\rvx - \rvx^{(i)})^\trns \mSigma^{-1} \frac{(\rvx^{(i)} - \rvx^{(j)})}{2}\\
    &\qquad \qquad \qquad \qquad \qquad \qquad \qquad + \frac{(\rvx^{(i)} - \rvx^{(j)})}{2}^\trns \mSigma^{-1} \frac{(\rvx^{(i)} - \rvx^{(j)})}{2} - \frac{(\rvx^{(i)} - \rvx^{(j)})}{2}^\trns \mSigma^{-1} \frac{(\rvx^{(i)} - \rvx^{(j)})}{2} \\
    &\qquad \qquad \qquad \qquad \qquad \qquad \qquad + \frac{1}{2}(\rvx^{(i)} - \rvx^{(j)})^\trns \mSigma^{-1} (\rvx^{(i)} - \rvx^{(j)}) \bigg) \ud \rvx\\
    &= \frac{1}{2} \frac{1}{(2\pi)^{D/2}\sigma_1^D \abs{\mSigma}^{1/2}} \int \exp \bigg(-\frac{1}{2\sigma_1^2}\bigg( (\rvx - \rvx^{(i)} + \frac{(\rvx^{(i)} - \rvx^{(j)})}{2})^\trns \mSigma^{-1} (\rvx - \rvx^{(i)} + \frac{(\rvx^{(i)} - \rvx^{(j)})}{2}) \\
    &\qquad \qquad \qquad \qquad \qquad \qquad \qquad + \frac{1}{4}(\rvx^{(i)} - \rvx^{(j)})^\trns \mSigma^{-1} (\rvx^{(i)} - \rvx^{(j)}) \bigg) \ud \rvx\\
    &= \frac{1}{2} \exp \bigg( -\frac{1}{8\sigma_1^2} (\rvx^{(i)} - \rvx^{(j)})^\trns \mSigma^{-1} (\rvx^{(i)} - \rvx^{(j)}) \bigg) \\
    &\qquad \qquad \int \frac{1}{(2\pi)^{D/2}\sigma_1^D \abs{\mSigma}^{1/2}} \exp \bigg(-\frac{1}{2\sigma_1^2}\bigg( (\rvx - \frac{(\rvx^{(i)} + \rvx^{(j)})}{2})^\trns \mSigma^{-1} (\rvx - \frac{(\rvx^{(i)} + \rvx^{(j)})}{2}) \bigg) \ud \rvx\\
    &= \frac{1}{2} \exp \bigg( -\frac{1}{8\sigma_1^2} (\rvx^{(i)} - \rvx^{(j)})^\trns \mSigma^{-1} (\rvx^{(i)} - \rvx^{(j)}) \bigg)
    \\
    &\implies \frac{1}{\sigma_1^2} (\rvx^{(i)} - \rvx^{(j)})^\trns \mSigma^{-1} (\rvx^{(i)} - \rvx^{(j)}) \approx 1 \\
    &\implies (\sqrt{\mSigma^{-1}} (\rvx^{(i)} - \rvx^{(j)}))^\trns (\sqrt{\mSigma^{-1}} (\rvx^{(i)} - \rvx^{(j)})) \approx \sigma_1^2 \\
    &\implies \Norm{ \sqrt{\mSigma^{-1}} (\rvx^{(i)} - \rvx^{(j)}) }_2 \approx \sigma_1 \\
    &\implies \Norm{ \sigma_N\ \Real \big( \rmW_N^{-1} \rmK \rmW_N (\rvx^{(i)} - \rvx^{(j)}) \big) }_2 \approx \sigma_1 \\
    &\implies \Norm{\sigma_N \rmW_N^{-1} \rmK \rmW_N \rvx^{(i)} - \sigma_N \rmW_N^{-1} \rmK \rmW_N \rvx^{(j)} }_2 \approx \sigma_{1}
\end{align*}

% For CIFAR10, this $\sigma_{1} \approx 3.72$.
For CIFAR10, this $\sigma_{1} \approx 20$ for NI-SMLD (whereas for SMLD $\sigma_1 \approx 50$).

\subsection{Other noise scales}

\begin{align*}
p_{\sigma_i}(r) &= \gN(r \mid m_i, s_i^2), \text{where } m_i \triangleq \sqrt{D}\sigma_i ; s_i^2 \triangleq \sigma_i^2 / 2\\
\gI_i &\triangleq (m_i - 3s_i, m_i + 3s_i)\\
p_{\sigma_i}(r \in \gI_{i-1})
&= \Phi\left( \frac{(m_{i-1} + 3s_{i-1}) - m_i}{s_i} \right) - \Phi\left( \frac{(m_{i-1} - 3s_{i-1}) - m_i}{s_i} \right)\\
&= \Phi\left(\frac{\sqrt{2}}{\sigma_i} (\sqrt{D}\sigma_{i-1} + \frac{3\sigma_{i-1}}{\sqrt{2}} - \sqrt{D}\sigma_i) \right) - \Phi\left(\frac{\sqrt{2}}{\sigma_i} (\sqrt{D}\sigma_{i-1} - \frac{3\sigma_{i-1}}{\sqrt{2}} - \sqrt{D}\sigma_i) \right) \\
&= \Phi\left(\frac{1}{\sigma_i} (\sqrt{2D}(\sigma_{i-1} - \sigma_i) + 3\sigma_{i-1}) \right) - \Phi\left(\frac{1}{\sigma_i} (\sqrt{2D}(\sigma_{i-1} - \sigma_i) - 3\sigma_{i-1}) \right) \\
&= \Phi\left( \sqrt{2D}(\gamma - 1) + 3\gamma \right) - \Phi \left( \sqrt{2D}(\gamma - 1) - 3\gamma \right) \approx 0.5
\end{align*}

Hence, the value of $\gamma$ remains (almost) the same:

% $\gamma = 1.0376858468648256$ ($\sigma_1=3.72, \sigma_L=0.01, L=161$).

$\gamma = 1.0375867506951884$ ($\sigma_1=20, \sigma_L=0.01, L=207$).

(wheras ealier $\gamma = 1.0375591319992028$ ($\sigma_1=50, \sigma_L=0.01, L=232$)

\subsection{Configuring annealed Langevin dynamics}

Let $\gamma = \frac{\sigma_{i-1}}{\sigma_i}$. For $\alpha = \epsilon\cdot  \frac{\sigma_i^2}{\sigma_L^2}$, we have $\rvx_T \sim \gN(\vzero, \Var[\rvx_T])$, where
\begin{align}
    \frac{\Var[\rvx_T]}{\sigma_i^2} &= \gamma^2 \rmP^T \mSigma \rmP^T + \frac{2\epsilon}{\sigma_L^2} \sum_{t=0}^{T-1} ( \rmP^t \mSigma \rmP^t)
\end{align}

\textbf{Proof:}

First, the conditions we know are
\begin{gather*}
\rvx_0 \sim p_{\sigma_{i-1}}(\rvx) = \gN(\rvx \mid \vzero, \sigma_{i-1}^2 \mSigma) = \frac{1}{(2\pi)^{D/2}\sigma_{i-1}^D\abs{\mSigma}^{1/2}} \exp \left( -\frac{1}{2\sigma_{i-1}^2} \rvx^\trns \mSigma^{-1} \rvx \right),\\
\nabla_\rvx \log p_{\sigma_i}(\rvx_t) = \nabla_\rvx  \left(-\log(\text{const.}) - \frac{1}{2\sigma_i^2} \rvx_t^\trns \mSigma^{-1} \rvx_t \right) = -\frac{1}{\sigma_i^2} \mSigma^{-1} \rvx_t, \\
\rvx_{t+1} \gets \rvx_{t} + \alpha \nabla_\rvx \log p_{\sigma_i}(\rvx_t) + \sqrt{2\alpha} \rvg_t = \rvx_t - \alpha \frac{1}{\sigma_i^2}\mSigma^{-1}\rvx_t + \sqrt{2\alpha} \rvg_t,
\end{gather*}
where $\rvg_t \sim \gN(\vzero, \mSigma)$, $\alpha = \epsilon \frac{\sigma_i^2}{\sigma_L^2}$. Therefore, the variance of $\rvx_t$ satisfies
\begin{align*}
    \Var[\rvx_t] = \begin{cases}
        \sigma_{i-1}^2 \mSigma \quad & \text{if $t = 0$}\\
        \Var[\big(\rmI - \frac{\alpha}{\sigma_i^2}\mSigma^{-1}\big) \rvx_{t-1}] + 2\alpha \mSigma \quad & \text{otherwise}.
    \end{cases}
\end{align*}

\begin{align*}
\Var[\rmA \rvx] &= \rmA \Var[\rvx] \rmA^\trns
% \implies \Var[\big(\rmI - \frac{\alpha}{\sigma_i^2}\mSigma^{-1}\big) \rvx_{t-1}]
% = \big(\rmI - \frac{\alpha}{\sigma_i^2}\mSigma^{-1}\big) \Var[\rvx_{t-1}] \big(\rmI - \frac{\alpha}{\sigma_i^2}\mSigma^{-1}\big)
\\
\implies \Var[\rvx_t]
&= \big(\rmI - \frac{\alpha}{\sigma_i^2}\mSigma^{-1}\big) \Var[\rvx_{t-1}] \big(\rmI - \frac{\alpha}{\sigma_i^2}\mSigma^{-1}\big) + 2\alpha \mSigma\\
\text{Let } \rmP = \rmI - \frac{\alpha}{\sigma_i^2}\mSigma^{-1} = \rmI - \frac{\epsilon}{\sigma_L^2}\mSigma^{-1} \\
\implies \Var[\rvx_t]
&= \rmP \Var[\rvx_{t-1}] \rmP + 2\alpha \mSigma\\
&= \rmP (\rmP \Var[\rvx_{t-2}] \rmP + 2\alpha \mSigma) \rmP + 2\alpha \mSigma\\
&= \rmP \rmP \Var[\rvx_{t-2}] \rmP \rmP + 2\alpha ( \rmP \mSigma \rmP + \mSigma)\\
&= \rmP^{(2)} \Var[\rvx_{t-2}] \rmP^{(2)} + 2\alpha ( \rmP \mSigma \rmP + \mSigma)\\
\implies \Var[\rvx_T]
&= \rmP^{(T)} \Var[\rvx_0] \rmP^{(T)} + 2\alpha \sum_{t=0}^{T-1} ( \rmP^{(t)} \mSigma \rmP^{(t)})\\
&= \sigma_{i-1}^2 \rmP^{(T)} \mSigma \rmP^{(T)} + 2\epsilon \frac{\sigma_i^2}{\sigma_L^2} \sum_{t=0}^{T-1} ( \rmP^{(t)} \mSigma \rmP^{(t)})\\
\implies \frac{\Var[\rvx_T]}{\sigma_i^2} &= \gamma^2 \rmP^{(T)} \mSigma \rmP^{(T)} + \frac{2\epsilon}{\sigma_L^2} \sum_{t=0}^{T-1} ( \rmP^{(t)} \mSigma \rmP^{(t)})
\end{align*}

Hence, we choose $\epsilon$ s.t. $\frac{Var[\rvx_T]}{\sigma_i^2} \approx 1$:

$\epsilon = 3.1\mathrm{e}{-7}$ for $T=5$, $\epsilon = 2.0\mathrm{e}{-6}$ for $T=1$

(whereas earlier $\epsilon = 6.2\mathrm{e}{-6}$ for $T=5$)

\subsection{Calculus of Variations}

Alexia et. al. discovered in Appendix E that the unconditional score model's estimate of the score in the case of a single data point $\rvx_0$ is:
\begin{align}
\rvs_\rvtheta(\tilde \rvx) &= \frac{1}{L}\sum_{i=1}^L \left( \frac{\EE_{\rvx \sim q_{\sigma_i}(\rvx \mid \tilde\rvx)}[\rvx] - \tilde\rvx}{\sigma_i} \right) = \frac{\rvx_0 - \tilde\rvx}{\sigma_H} \\
\rvs_\rvtheta(\tilde \rvx, \sigma_i) &= \frac{1}{\sigma_i}\rvs_\rvtheta(\tilde \rvx) = \frac{\rvx_0 - \tilde\rvx}{\sigma_i \sigma_H} \neq \frac{\rvx_0 - \tilde\rvx}{\sigma_i^2} 
\end{align}
where $\frac{1}{\sigma_H} = \frac{1}{L}\sum_{i=1}^L \frac{1}{\sigma_i}$, i.e. $\sigma_H$ is the harmonic mean of the $\sigma_i$s used to train.

In our case,
\begin{align*}
\frac{\partial \gL_2}{\partial \rvs} &= \int \int q_{\sigma}(\tilde\rvx, \rvx, \sigma) \left( \rvs(\tilde\rvx) + \sqrt{\mSigma^{-1}}\frac{\tilde\rvx - \rvx}{\sigma} \right) \D\rvx \D \sigma = 0 \\
&\iff \rvs(\tilde\rvx)q(\tilde\rvx) = \sqrt{\mSigma^{-1}} \int \int q_{\sigma}(\tilde\rvx, \rvx)p(\sigma) \left( \frac{\tilde\rvx - \rvx}{\sigma} \right) \D\rvx \D \sigma \\
&\iff \rvs(\tilde\rvx)q(\tilde\rvx) = \sqrt{\mSigma^{-1}} \EE_{\sigma \sim p(\sigma)} \left[ \int q_{\sigma}(\tilde\rvx \mid \rvx) q(\tilde\rvx) \left( \frac{\tilde\rvx - \rvx}{\sigma} \right) \D\rvx \right] \\
&\iff \rvs(\tilde\rvx) = \sqrt{\mSigma^{-1}} \EE_{\sigma \sim p(\sigma)} \left[ \int q_{\sigma}(\tilde\rvx \mid \rvx) \left( \frac{\tilde\rvx - \rvx}{\sigma} \right) \D\rvx \right] \\
&\iff \rvs(\tilde\rvx) = \sqrt{\mSigma^{-1}} \EE_{\sigma \sim p(\sigma)} \left[ \frac{\EE_{\rvx \sim q_\sigma(\rvx \mid \tilde\rvx)}[\rvx] - \rvx}{\sigma} \right]
\end{align*}
In the case of a single data point $\rvx_0$:
\begin{align*}
\rvs(\tilde\rvx) &= \sqrt{\mSigma^{-1}} \frac{\rvx_0 - \rvx}{\sigma_H} \\
\rvs_\rvtheta(\tilde \rvx, \sigma_i) &= \frac{1}{\sigma_i}\sqrt{\mSigma^{-1}}\rvs_\rvtheta(\tilde \rvx) = \mSigma^{-1} \frac{\rvx_0 - \tilde\rvx}{\sigma_i \sigma_H} \neq \mSigma^{-1} \frac{\rvx_0 - \tilde\rvx}{\sigma_i^2}
\numberthis
\end{align*}

Hence, we correct for it while sampling:
\begin{align}
\rvs_\rvtheta(\tilde\rvx, \sigma_i) = \frac{\sigma_H}{\sigma_i^2} \sqrt{\mSigma^{-1}} \rvs_\rvtheta(\tilde\rvx) = \frac{\sigma_H \sqrt{\Var[\mSigma_N]}}{\sigma_i^2} \Real(\rmW_N^{-1} \rmK \rmW_N \rvs_\rvtheta(\tilde\rvx))
\end{align}

\subsection{beta for CAS}

\begin{align*}
\rvx_{t+1} &\leftarrow \rvx_t + \eta \sigma_t^2 s^*( \rvx_t, \sigma_t) + \sigma_{t+1} \beta \rvg \qquad (\eta = \eps/\sigma_L^2) \\
\implies \rvx_{t+1} &\leftarrow \rvx_t + \eta \mSigma^{-1}(EDS - \rvx_t) + \sigma_{t+1} \beta \rvg \\
\implies \rvx_{t+1} &\leftarrow (\rmI - \eta \mSigma^{-1})\rvx_t + \eta \mSigma^{-1}EDS + \sigma_{t+1} \beta \rvg
\end{align*}

Noise component is:
\begin{align*}
(\rmI - \eta \mSigma^{-1})\sigma_t \rvg_t + \sigma_{t+1} \beta \rvg
\end{align*}

Variance of noise component is:
\begin{align*}
&\sigma_t^2 (\rmI - \eta \mSigma^{-1}) \Var[\rvg_t](\rmI - \eta \mSigma^{-1})^\trns + \sigma_{t+1}^2 \beta^2 \Var[\rvg]
\end{align*}

Each diagonal element of Variance of noise component is:
\begin{align*}
&\sigma_t^2 (1 - \eta)^2 + \sigma_{t+1}^2 \beta^2 = \sigma_{t+1}^2 \left(\gamma^2(1 - \eta)^2 + \beta^2\right) = \sigma_{t+1}^2 \\
\implies &\beta = \sqrt{1 - \gamma^2(1 - \eta)^2} \qquad \text{where } \gamma = \sigma_t/\sigma_{t+1}
\end{align*}

In this case, the signal-to-noise ratio is:
\begin{align*}
\EE\left[ \Norm{ \frac{\eta \sigma_t^2 \rvs_\rvtheta(\rvx_t, \sigma_t)}{\sigma_{t+1} \beta \rvg} }_2 \right] = \frac{\eta \gamma \sigma_t}{\beta} \frac{\EE\left[ \Norm{ \rvs_\rvtheta(\rvx_t, \sigma_t) }_2 \right]}{\EE\left[ \Norm{ \rvg }_2 \right]} \approx \frac{\eta \gamma}{\beta}
\end{align*}

\end{document}